\documentclass[10pt,journal,compsoc]{IEEEtran}

\usepackage{times}
\usepackage{epsfig}
\usepackage{graphicx}
\usepackage{amsmath}
\usepackage{amssymb}
\usepackage{dutchcal}
\usepackage{algorithm}
\usepackage{algorithmic}
\usepackage{mathtools}
\usepackage{array}
\usepackage{multirow}
\usepackage[inline]{enumitem}
\usepackage{subfig}
\usepackage[pagebackref=false,breaklinks=true,letterpaper=true,allcolors=blue,colorlinks,bookmarks=false]{hyperref}

\ifCLASSOPTIONcompsoc
  \usepackage[nocompress]{cite}
\else
  \usepackage{cite}
\fi
\ifCLASSINFOpdf
\else
\fi
\begin{document}
%
\title{Discriminatively Trained Latent Ordinal Model \\ for Video Classification}
%
%
%
%

\author{Karan~Sikka,~\IEEEmembership{Member,~IEEE,}
        and~Gaurav~Sharma,~\IEEEmembership{Member~IEEE}
\IEEEcompsocitemizethanks{\IEEEcompsocthanksitem Karan Sikka is with SRI
	International, Princeton. Part of this work was done when he was at University of California San Diego.
\texttt{ksikka.com}
\IEEEcompsocthanksitem Gaurav Sharma is with the Indian Institute of Technology Kanpur,
India. Part of the work was done when he was with the Max Planck Institute for Informatics,
Germany. \texttt{http://www.grvsharma.com}
}
}

\def\etal{et al.\ }
\def\etc{etc.\ }
\def \ie {i.e.\ }
\def \eg {e.g.\ }
\def\cf{cf.\ }
\def\vs{vs.\ }
\def\pd{\partial}
\def\grad{\nabla}
\def\Li{\mathcal{L}}
\def\O{\mathcal{O}}
\def\N{\mathbb{N}}
\def\c{\mathbcal{c}}
\def\Lt{\tilde{L}}
\def\R{\mathbb{R}}
\def\X{\mathcal{X}}
\def\I{\mathcal{I}}
\def\F{\mathcal{F}}
\def\L{\mathcal{L}}
\def\w{\textbf{w}}
\def\x{\textbf{x}}
\def\k{\textbf{k}}
\def\t{\textbf{t}}
\def\d{\boldsymbol{\delta}}
\def\xib{\boldsymbol{\xi}}
\def\y{\textbf{y}}
\def\l{\boldsymbol{\ell}}
\def\wrt{w.r.t.\ }
\def\a{\boldsymbol{\alpha}}
\def\vertspace{0.6em}

\newcommand{\red}[1]{\textcolor{red}{#1}}
\def\algorithmautorefname{Algorithm}
\def\figureautorefname{Fig.}
\def\tableautorefname{Tab.}
\def\equationautorefname{Eq.}
\def\sectionautorefname{Sec.}
\def\subsectionautorefname{Sec.}
\def\subsubsectionautorefname{Sec.}

\IEEEtitleabstractindextext{%

\begin{abstract}
We address the problem of video classification for facial analysis and human action recognition. We
propose a novel weakly supervised learning method that models the video as a sequence of
automatically mined, discriminative sub-events (\eg onset and offset phase for ``smile", running and
jumping for ``highjump''). The proposed model is inspired by the recent works on Multiple Instance
Learning and latent SVM/HCRF -- it extends such frameworks to model the ordinal aspect
in the videos, approximately. We obtain consistent improvements over relevant
competitive baselines on four challenging and publicly available video based facial analysis
datasets for prediction of expression, clinical pain and intent in dyadic conversations, and on three
challenging human action datasets. We also validate the method with qualitative results and show
that they largely support the intuitions behind the method.
\end{abstract}

}

\maketitle

\IEEEdisplaynontitleabstractindextext

%
\IEEEpeerreviewmaketitle

\section{Introduction}
\label{intro}

\IEEEPARstart{T}{he} world is exploding with large amounts of visual data and videos are the major
chunk of such data. Hundreds of hours of videos are being added to YouTube per day, and millions of
surveillance cameras are recording continuous video streams day in and day out. Most of these videos
are human centric and the huge scale demands systems capable of automatically processing and understanding
such data. Hence, researchers in computer vision have been actively engaged in designing methods
which work with human centered video data, for tasks like recognizing human actions in videos
\cite{laptevCVPR2008, gaidonPAMI2013, wangIJCV2015, SimonyanNIPS2014, kuehneICCV2011,
niebles2010modeling, scovanner2007, karCVPR2017} and facial analysis in videos
\cite{sikka2014classification, lucey2011painful, viola2006multiple, rudovic2012multi,
simonCVPR2010, sharmaCVPRW15}. 

We are interested in the challenging and relevant problem of classifying videos, of faces and
humans, based on the property the face is exhibiting or the actions the human is performing. We work
in a weakly supervised setting where only labels for the videos are given while all individual, or
any selected subset of, frames do not have any labels. The usual assumption in such weakly supervised
setting is that the positive `bag' contains at least one positive instance, while the negative `bag'
does not have any positive instances. In weakly supervised settings, Multiple Instance Learning
(MIL) \cite{andrews2002support} based methods are one of the popular approaches and have been
applied to the task of facial video analysis \cite{sikka2014classification, ruiz2014regularized,
Wu2015FG} with video level, and no frame level, annotations. However, the main drawbacks of most of
such (MIL based) approaches are that (i) they use the maximum scoring vector to make the prediction
\cite{andrews2002support}, and (ii) the temporal/ordinal information is lost completely. While, in
the recent work by Li and Vasconcelos \cite{liCVPR2015}, MIL framework has been extended to consider
multiple top scoring vectors, the temporal order is still not incorporated. Intuitively, the
temporal order is definitely important; recent works exploit it for related computer vision tasks,
\eg visual representation learning \cite{FernandoCVPR2017, LeeICCV2017}. We propose a novel method
that (i) works with weakly supervised data, (ii) mines out the prototypical and discriminative set
of vectors required for the task, and (iii) learns constraints on the temporal order of such
vectors. We show how modeling multiple vectors instead of the maximum one, while simultaneously
considering their ordering, leads to improvements in performance. 

The proposed model belongs to the family of models with structured latent variables \eg Deformable
Part Models (DPM) \cite{felzenszwalbPAMI2010} and Hidden Conditional Random Fields (HCRF)
\cite{wang2009max}. In DPM, Felzenszwalb \etal \cite{felzenszwalbPAMI2010} constrain the location of
the parts (latent variables) to be around fixed anchor points with penalty for deviation while Wang
and Mori \cite{wang2009max} impose a tree structure on the human parts (latent variables) in their
HCRF based formulation. In contrast, we are not interested in constraining our latent variables
based on fixed anchors \cite{felzenszwalbPAMI2010, niebles2010modeling} or distance (or correlation) among themselves
\cite{wang2009max,rudovic2012multi}, but are only interested in modeling the order in which they
appear. Thus, the model is stronger than models without any structure while being weaker than models
with more strict structure \cite{felzenszwalbPAMI2010, wang2009max, niebles2010modeling}.

The proposed model is also reminiscent of Actom Sequence Model (ASM) of Gaidon \etal
\cite{gaidonPAMI2013}, where a temporally ordered sequence of sub-events are used to perform action
recognition in videos.  However, while ASM requires annotation of such sub-events in the videos; the
proposed model aims to find such sub-events automatically. While ASM places absolute temporal
localization constraints on the sub-events, the proposed model only cares about the order in which
such sub-events occur. One advantage of doing so is the flexibility of sharing appearances for two
sub-events, especially when they are automatically mined. As an example, the facial expression may
start, as well as end, with a neutral face. In such case, if the sub-event (neutral face) is tied to
a temporal location we will need two redundant (in appearance) sub-events \ie one at the beginning
and one at the end. While, here such sub-events will merge to a single appearance model, with the
symmetry encoded with similar cost for the two ordering of such sub-event, keeping the rest same.

Informally,  the proposed model is a collection of discriminative templates, which capture the
appearances of the sub-events in the video, along with a cost vector corresponding to all possible
permutations in which the events can occur in a video. Scoring by the proposed model is done as
follows.  The multiple discriminative events in the model are detected and scored in the current
video and they also incur a cost depending on the temporal ordering in which they appear in the
video. The mining of such discriminative events and learning of the appropriate templates, along
with learning the costs associated with different orders of occurrence of the events, happens fully
automatically in a weakly supervised setting with the event locations (in time) being latent
variables. We propose to learn the model with a max-margin loss minimization objective, optimized
with efficient stochastic gradient descent. On the task of facial video analysis, we validate the
model on four challenging datasets of expression recognition (CK+ \cite{cohnkanade} and Oulu-CASIA
VIS \cite{zhao2011facial} datasets), clinical pain prediction (UNBC-McMaster Pain dataset
\cite{lucey2011painful}) and intent prediction in dyadic conversations (LILiR dataset
\cite{sheerman2009feature}). We show that the method consistently outperforms temporal pooling and
MIL based competitive baselines. In combination with complementary features, we report
state-of-the-art results on these datasets with the proposed model.  

On the task of human video analysis, we follow previous work and propose a second variant of the
method which takes into account the global temporal information as well. The model with only local
discriminative sub-events assumes that the video can be factorized cleanly into such sub-events.
However, for the more challenging case of human action such factorization is not always clean, in
the sense that while some actions are easily factorizable, \eg running and jumping for high-jump,
others are a complex combination of local events and full global temporal information and context,
\eg hitting.  This interplay of local and global factors for action recognition has been
acknowledged either explicitly or implicitly by experiments in previous works as well
\cite{niebles2010modeling, tang2012learning}. We take inspiration from such works and extend the
model to incorporate global features, obtained by some pooling operation over the features of the
frames, of the video as well. We cast the final objective as a weighted (convex) combination of the
local and global parts and learn the parameters jointly. We validate the model, for human analysis,
on challenging datasets of Olympic Sports \cite{niebles2010modeling}, Human Motion Database (HMDB)
\cite{kuehneICCV2011} and HighFive dataset \cite{patron2010high} of human interactions. On human
analysis as well, we show consistent improvements over challenging and relevant baselines. The
method achieves results that are competitive to state-of-the-art on the three datasets. Further, qualitative
analysis of the results validate the hypothesis of the method \ie we show that the method is
successful in mining out discriminative events and also learns a consistent ordering over their
occurrences.

A preliminary version of this work appeared in \cite{SikkaCVPR2016} with the first variant of the
model applied only to facial analysis tasks.

\section{Related Works}	

We now describe closely related works in the following sections. We compare and contrast our model with
related models in the literature while discussing works on the tasks of face and general
human action video classification.

\subsection{Facial Analysis}
Facial analysis is an important area of computer vision. The representative problems include face
(identity) recognition \cite{Parkhi15}, identity based face pair matching \cite{LFWtech},
age estimation \cite{alnajar2014expression, BhattaraiICASSP2016}, kinship
verification \cite{lu2014neighborhood}, emotion prediction \cite{fasel2003automatic},
\cite{kaltwang2012continuous}, among others.  Facial analysis finds important and relevant real
world applications such as human computer interaction, personal robotics, and patient care in
hospitals \cite{sikka2014classification, lucey2011painful, viola2006multiple, de2011facial}. When we
work with videos of faces, we assume that face detection has been done reliably. We note that,
despite reduction to just faces, the problem is still quite challenging due to variations in human
faces, articulations, lighting conditions, poses, video artifacts such as blur \etc Moreover, we
work in a weakly supervised setting, where only video level annotation is available and there are no
annotations for individual video frames. 

Early approaches for facial expression recognition used apex (maximum expression) frames
\cite{sikka2012, ojala2002multiresolution, de2011facial} or pre-segmented clips, and thus were
strongly supervised. Also, they were often evaluated on posed video datasets \cite{cohnkanade}. To
encode the faces into numerical vectors, many successful features were proposed \eg Gabor
\cite{littlewort2011computer} and Local Binary Patterns (LBP) \cite{ojala2002multiresolution},
fiducial points based descriptors \cite{zhang1998comparison}. They handled videos by either
aggregating features over all frames, using average or max-pooling \cite{laptevCVPR2008,
sikkaexemplar}, or extending features to be spatio-temporal \eg 3D Gabor \cite{wu2010facial} and
LBPTOP \cite{zhao2007dynamic}.  Facial Action Units, representing movement of facial muscle(s)
\cite{de2011facial}, were automatically detected and used as high level features for video
prediction \cite{de2011facial, littlewort2011motion}.

Noting that temporal dynamics are important for expressions \cite{de2011facial}, the recent focus
has been more on algorithms capturing dynamics \eg Hidden Markov Model (HMM) \cite{
lien1998automated} and Hidden Conditional Random Fields (HCRF) \cite{chang2009learning,
quattoni2007hidden} have been used for predicting expressions. Chang \etal
\cite{chang2009learning} proposed a HCRF based model that included a partially observed hidden state
at the apex frame, to learn a more interpretable model where hidden states had specific meaning.
The models based on HCRF are also similar to latent structural SVMs \cite{wang2009max,
simonCVPR2010}, where the structure is defined as a linear chain over the video frames. Other
discriminative methods were proposed based on Dynamic Bayesian Networks \cite{zhang2005active} or
hybrids of HMM and SVM \cite{valstar2012fully}. Lorincz \etal \cite{lorincz2013emotional} explored
time-series kernels \eg based on Dynamic Time Warping (DTW) for comparing expressions. Another
similar model used probabalistic kernels for classifying exemplar HMM models \cite{sikkaexemplar}. 

Nguyen \etal \cite{nguyen2009weakly} proposed a latent SVM based algorithm for classifying and
localizing events in a time-series. They later proposed a fully supervised structured SVM for
predicting Action Unit segments in video sequences \cite{simonCVPR2010}. Our algorithm differs
from \cite{nguyen2009weakly}, while they use simple MIL, we detect multiple prototypical segments
and further learn their temporal ordering. MIL based algorithm has also been used for predicting
pain \cite{sikka2014classification}. In recent works, MIL has been used with HMM \cite{Wu2015FG} and
also to learn embedding for multiple concepts \cite{ruiz2014regularized} for predicting facial
expressions. Rudovic \etal \cite{rudovic2012multi} proposed a CRF based model that accounted for
ordinal relationships between expression intensities. Our work differs from this work in handling
weakly labeled data and modeling the ordinal sequence between sub-events.

\subsection{Human Analysis} 

There have been many works related to understanding humans in visual data.  Methods have been
proposed for human action and attribute recognition from still images \cite{sharmaTPAMI2017, bourdevICCV2011,
bourdevICCV2009, majiCVPR2011, gkioxariICCV2015, sharmaBMVC2011,
sharmaCVPR2012}, where typical appearances (\eg sports clothes) or poses (\eg
jumping, riding a bike) may be sufficient for recognition \cite{kuehneICCV11}. While many actions
are highly correlated with typical poses and clothes, many require some temporal information and
using just still images are thus not sufficient. Motion, being an important cue for the task of human
action recognition, has been exploited by many works \cite{laptevCVPR2008, gaidonPAMI2013,
wangIJCV2015, SimonyanNIPS2014, kuehneICCV11, zhuICCV2013, wuCVPR2014, wangICCV2013}.  One of the
most popular approaches in the last decade relied on using local methods that extracted descriptors
such Histogram of Flow (HOG) and Histograms of Gradients (HOF) over salient 3D regions such as
Space-Time Interest Points \cite{laptevCVPR2008, dollar2005behavior, wang2009evaluation}.  The final
descriptors were obtained by popular pooling methods such as the bag-of-features
\cite{CsurkaSLCV2004, wang2009evaluation} and Fisher vector \cite{perronninECCV2010,
wang2012comparative} frameworks.  Since pixels are moving over time in videos, a fixed space-time
voxel may not be able to represent complete motion of such a pixel. Therefore techniques were
proposed to describe trajectories by tracking pixel(s) over time instead of interest points
\cite{wang2013dense, sun2010activity, wangIJCV2015, messing2009activity, jain2013representing}. Wang
\etal \cite{wang2013dense} showed that densely tracking motion trajectories is superior to previous
approaches and proposed to describe them using Motion Boundary Histogram (MBH) for handling camera
motion. Several approaches improved upon the Dense Trajectories approach by compensating for camera
motion by removing background motion using affine transformation \cite{jain2013representing},
clustering trajectories for identifying dominant camera motion \cite{jiang2012trajectory}, using
image-stitching methods to generate stabilized video prior to computing trajectories
\cite{gaidon2014activity}. Wang \etal proposed Improved Dense Trajectories (iDT)
\cite{wangIJCV2015} approach that estimated and removed camera motion using homography and also
removed
inconsistent feature matches by human detection. In combination with using Histogram based features,
Fisher Vector encoding and Spatio-Temporal pyramids, they showed significant improvements compared
to previous methods. \cite{lan2015beyond} improved upon these features by proposing to stack
features extracted at multiple temporal frequencies.  

Many works have also focussed on improving upon standard pooling pipelines by explicitly modeling
spatial and temporal structure of human activities \cite{niebles2010modeling, tang2012learning,
gaidon2014activity, li2013dynamic, brendel2011learning, raptis2013poselet, bilen2011object}. Niebles
\etal \cite{niebles2010modeling} used a variant of DPM approach to model each activity as composed
of short temporal segments with anchored locations, and penalized segments that drifted away from
these anchors during inference. Improving upon their approach, Tang \etal \cite{tang2012learning}
proposed a more flexible variable duration Hidden Markov model that modeled an activity as composed
of temporal segments with variable durations and first order transitions. Some works have used MIL
based approaches for action classification \cite{bilen2011object, satkin2010modeling,
nguyen2009weakly}. Li \etal \cite{li2013dynamic} proposed to dynamically pool over relevant
segments in a video. A closely related work to ours modeled videos as a set of sparse key-frames,
that were learned weakly, but assumed fixed ordering between the events \cite{raptis2013poselet}.
Gaidon \etal \cite{gaidonPAMI2013} used additional training data for learning the sub-events while assuming fixed
ordering. In constrast we not only model each video as composed of sub-events but also learn them in
a weakly supervised setting along with a loose ordering between them.  Moving away from latent
variable modeling, Gaidon \etal \cite{gaidon2014activity} used a tree based kernel to compare
hierarchical decomposition of videos as cluster of dense trajectories. Other approaches have used
attribute dynamics \cite{liu2011recognizing}, mid-level parts \cite{ma2015space}, spatio-temporal
graphs \cite{brendel2011learning}, higher level pooling methods such as rank pooling
\cite{fernando2015modeling} and distribution of classifier scores \cite{hoai2014improving}.

As deep learning showed excellent performance for image classification problems
\cite{krizhevskyNIPS2012, SimonyanICLR2015, he2015deep}, several approaches also explored their use
for human action recognition \cite{tran2015learning, ballas2015delving, li2016videolstm,
SimonyanNIPS2014, rohrbach2012database, feichtenhofer2016convolutional, zhu2016key}. Some earlier
works relied on using 3D convolutional networks \cite{ji20133d} and their extensions such as 
early and late fusion \cite{KarpathyCVPR2014} for action recognition. Simonyan \etal
\cite{SimonyanNIPS2014} proposed  a two-stream convolutional network that learned both spatial and
temporal networks (over stacked optical flow frames). A drawback of these networks was that they
were learning features over a few frames and  thus were only marginally better than networks that
learned features over a single frame.  Ng \etal \cite{YueCVPR2015} proposed a network that allows training over longer
periods by either using feature pooling (similar to temporal pooling) or Recurrent Neural Networks
such as Long Short Term Memories (LSTMs). Several other approaches have also explored the use of LSTMs architectures
\cite{ballas2015delving, sharma2015action} for action classification and
video caption generation \cite{donahueCVPR2015}. Some recent methods have also used Attention based
LSTMs that does classification while focussing on discriminative parts of a video
\cite{li2016videolstm, sharma2015action}. Tran \etal \cite{tran2015learning} showed that strong 3D
and compact spatiotemporal features can be learned for action recognition by using small 3D filters
of $3 \times 3 \times 3$ pixels. Several of these works have also shown the advantages of fusing deep models
trained on spatial and temporal components with non-deep features such as iDT
\cite{li2016videolstm}.

\section{Approach: Latent Ordinal Model (LOMo)}
\label{secApproach}
We denote a video as sequence of $N$ frames\footnote{We assume, for brevity, all videos have the same
number of frames, extension to different number of frames is immediate} represented as a matrix 
\begin{align}
X = [\x_1, \x_2, \ldots, \x_N]
\end{align}
with $\x_f \in \R^d$ being the feature vector for frame $f$. We work in
a weakly supervised binary classification setting, where we are given a training set 
\begin{align}
\X = \{(X,y)\} \subset \R^{d\times N} \times \{-1,+1\}
\end{align}
containing videos annotated with the presence ($y=+1$) or absence ($y=-1$) of a class in $X$,
without any annotations for specific frames of the video $X$ \ie $\x_f \forall f\in[1,N]$. While we
confine ourselves to the task of video classification in this paper, we note that our model
is applicable to general vector sequence classification in a weakly supervised setting. 

The proposed model learns a set of events, as well as a cost function associated with the order of
occurrence of those events. 
The events are defined by the associated,
discriminatively learned, templates. These templates capture the appearances of
different sub-events, \eg neutral, onset or offset phase of an expression, while the cost function
captures the discriminative likelihood of the different temporal orders in which the sub-events
appear in the videos. The model templates and the cost function are all automatically and jointly
learned, from the training data. Hence, the sub-events are not constrained to be either similar or
distinct \wrt each other, and are not fixed according to certain expected states \cite{gaidonPAMI2013}. They are mined from the data and
could potentially be a combination of the sub-events generally used by humans, \eg to describe expressions or
human activities \cite{gaidonPAMI2013, niebles2010modeling}. 

Formally, the model is defined as 
\begin{align}
        \Theta = \left(\{\w_i\}_{i=1}^M, \{c_j\}_{j=1}^{M!}\right), \w_i \in \R^d, c_j \in \R
\end{align}

with $i=1,\ldots,M$ indexing over the $M$ sub-event templates and $j=1,\ldots,M!$ indexing over the
different temporal orders in which these templates can occur.  The $\w_i$ are similar to the SVM
hyperplane parameter vectors, which have been often visualized as templates
\cite{felzenszwalbPAMI2010}. While, the ordering function is implemented as a look-up table \ie
$\c:\{1,\ldots, M!\}\rightarrow \R$ with $\c(i) = c_i \in \R$, with size equal to the number of
permutations of the sub-events.  In the following sections we make the model, and especially the
cost function, more concrete and give two variants of the proposed model adapted to facial analysis
and human action classification, respectively. We first describe the model that takes into account
the contribution of only the local temporal information that includes scores from the detected
sub-events and the ordering cost function (\autoref{secModellomo}). This model works well for
analyzing human facial behavior sequences that have a strong local temporal structure. However, this
might not be the case with unconstrained human activities where certain classes can be better
analyzed by either local or global motion components or their combinations. Thus, we adapt the model to
include both the local and global temporal information, that are adaptively weighted and learned for
different classes (\autoref{secModelAlomo}). 

We learn the model $\Theta$ with a $\ell_2$ regularized max-margin hinge loss minimization, given by 
\begin{align}
\label{eqnObj}
\min_{\Theta} \L(\Theta) = & \ \ \left\{ \frac{\lambda_1}{2} \|\w\|^2 + 
\frac{\lambda_2}{2} \sum_{j=1}^{M!} c_j^2 \right. \nonumber \\
             & \hspace{4em} \left. + \ \frac{1}{|\X|} \sum_{X \in \X} \left[1 - y_i s_\Theta(X) \right]_+
             \right\},
\end{align}
\begin{align}
& [a]_+ = \max(a,0)\ \forall a \in \R.
\end{align}
$s_\Theta(X)$ is the scoring function which uses the model
templates and the cost function to assign a confidence score to the example $X$ and $\w =
[\w_1,\ldots,\w_M]$ is the concatenation of the component
vectors. The decision boundary is given by $s_\Theta(X)=0$. The
scoring function depends on the type of model we use and is thus defined in the following sections
along with the two model variants. 

\subsection{Scoring Function with Local Events --- LOMo}
\label{secModellomo}
In the simplest case the model factorizes the sequence as ordered set of local events, which are
relatively few compared to the number of elements in the sequence. The model consists of a set of
sub-event templates and a function which assigns weights to all the different possible ordering of
the events. 

Deviating from a linear SVM classifier, which has a single parameter vector, the model now has multiple
such vectors which act at different temporal positions. The scoring function for a video $X$, with model
$\Theta$, is defined as 
\begin{subequations}
\begin{align}
s_\Theta(X) & = \max_\k   \frac{1}{M} \sum_{i=1}^M \w_i^\top \x_{k_i}
+ c_{\sigma(\k)} \\
& \textrm{\hspace{2.5em} s.t. \ } \mathcal{O}_v(\k) \leq \beta
\end{align}
\label{eqnScore1} 
\end{subequations}
where, 
\begin{align}
\k = [k_1,\ldots,k_M] \in \N^M, 1 \leq k_j \leq N
\end{align}
are the $M$ latent variables, and 
\begin{align}
\sigma : \N^M \rightarrow \N
\end{align}
maps $\k = (k_1,\ldots,k_M)$ to an index, with lexicographical ordering \eg
with $M=4$ and without loss of generality $k_1<k_2<k_3<k_4$, $\sigma(k_1,k_2,k_3,k_4) = 1,
\sigma(k_1,k_2,k_4,k_3) = 2, \sigma(k_1,k_3,k_2,k_4) = 3$ and so on. The latent variables take the
values of the frames on which the corresponding sub-event templates in the model gives maximal
response while being penalized by the cost function for the sequence of occurrence of the
sub-events. $\mathcal{O}_v(\k)$ is an overlap function, with $\beta$ being a threshold, to ensure that multiple
$\w_i$'s do not select very close by frames. We realize the overlap function by constraining the
temporal locations of different sub-events to occur at a certain distance from each other (refer to
\autoref{learning}).  

\subsection{Scoring Function with Local Events and Global Information --- Adaptive LOMo}
\label{secModelAlomo}

The LOMo model works well for facial videos, \eg of facial expressions, as they are expected to be
composed of temporal segments such as neutral, onset and apex states for facial expressions.
However, this is not always the case for human activities which are more complex and are both
spatially and temporally unconstrained. These activities could span from simple actions, that are
composed of a single motion segment, to complex or periodic activities such as ``talk'', ``longjump'', ``hugging'',
``highjump'' \etc In other words different activities have different temporal structure.  We adapt
LOMo to wide range of activities by extending the
model to include global information. We obtain global temporal
information by using temporal pooling over all the features of a video. This has
been often done in works learning decomposable temporal or spatial structure in videos
\cite{niebles2010modeling, tang2012learning, pandey2011scene, bilen2012classification}.  Going a step further, compared to such works,
we optimize an objective which is a convex combination of scorings based on local and global
components, jointly over the parameters corresponding to the two components. A similar idea was 
used by authors in \cite{gaidon2014activity}, who also did a weighted combination of local and global
components.  However, (i) the issue of weakly labelled data was not addressed and (ii) the locality
was defined by clustering local spatio-temporal features rather than being learned jointly with the
classifier, as in the present case.  

The adapted scoring function includes weighted combination of both, the local sub-events and their 
ordering cost and the global temporal component. The scoring function we use is given as,
\begin{subequations}
\begin{align}
s_\Theta(X) = 
              \max_\k & \left\{ \gamma_g \w_g^\top \x_g
             + \bar{\gamma_g} \left( \frac{\sum_{i=1}^M \w_i^\top \x_{k_i}}{M}
                + c_{\sigma(\k)}
        \right) \right\}\\
	& \textrm{\hspace{2.5em} s.t. \ } \mathcal{O}_v(\k) \leq \beta
\end{align}
\label{eqnScore2} 
\end{subequations}
where, in addition to the notations introduced above in \autoref{secModellomo}, $\bar{\gamma_g} = 1
- \gamma_g$ and 
\begin{align}
\x_g = \textrm{Pool}(X) \in \R^d
\end{align}
is the global feature, obtained by pooling over all the vectors for the different frames. $\w_g \in
\R^d $ is the hyperplane for the global template. 

In addition to the change in the scoring function, the objective function (\autoref{eqnObj}) is also
modified slightly by including $\ell_2$ regularization term for the global parameter $\w_g$, \ie the
$\w$ in \autoref{eqnObj} becomes $\w = [\w_g,\w_1,\ldots,\w_M]$.

\subsection{Discussion} 
Intuitively, in the proposed ordinal model, we capture the idea that each video sequence is composed
of a small number of prototypical sub-events, \eg onset followed by apex phase for a
smiling face and running followed by jumping for long jump. The components in our model
capture the appearance of the prototypical sub-events of the class of interest. However, instead
of the sub-events being manually defined, (i) they are learned within a
discriminative framework and (ii) are, thus, mined automatically with a discriminative objective.
The cost function $\c(\cdot)$ effectively learns the order in which such appearances should occur.
It is expected to support the likely order of sub-events while penalizing the unlikely ones. Even if
a negative video gives reasonable sub-event detections, the order of occurence of such false
positive detections is expected to be incorrect. Thus, the negative video is expected to be
penalized by the order dependent cost despite giving sub-event detections. We validate these 
intuitions empirically with qualitative results in \autoref{secQualRes}. 

The second variant of the model, described in \autoref{secModelAlomo}, combines local events with the global
information in the video using a convex combination of the two respective terms as the optimization
objective. This formulation adapts LOMo to different human action classes that could have either
local or global structure, or a combination of both. The relative importance of the local versus global
parts are learned using cross-validation for each class. We later discuss and show qualitative examples, in
\autoref{secQualRes}, of classes where either of the two components are important compared to the
other for human actions. The Adaptive LOMo is an extension of LOMo, setting $\gamma=0$ makes it same as 
to LOMo while $\gamma=1$ makes it same as the global temporal pooling based methods. 

\begin{algorithm}[t]
\begin{algorithmic}[1]
\STATE \emph{Given}: $\X, M, \lambda, \eta, k, \gamma_g$
\STATE \emph{Initalize}: $\w_i \sim U[0, 1e^{-4}] \forall i \in [1,M], \mathbf{c} \leftarrow \mathbf{0}$
\FORALL{$t = 1,\ldots,$ \texttt{maxiter}}
    \STATE Randomly sample $(X,y) \in \X$ 
    \STATE Obtain $s_\Theta(X)$ and $\k$ using \autoref{eqnScore1} or \autoref{eqnScore2}
    \IF{$ys_\Theta(X) < 1$}
        \FORALL{$i = 1,\ldots,M$}
            \STATE $\w_i \leftarrow \w_i(1-\lambda_1 \eta) + \eta (1-\gamma_g) y_i\x_{k_i}/M$  
            \label{update_eq}
        \ENDFOR \\
	\STATE $c \leftarrow c(1  - \lambda_2 \eta) $ \\
	\STATE $c_{\sigma(\k)} \leftarrow c_{\sigma(\k)} + \eta (1 - \gamma_g) y_i$ \\
        \STATE $\w_g \leftarrow \w_g(1-\lambda_1 \eta) + \eta \gamma_g y_i\x_g$  
    \ENDIF
\ENDFOR
\STATE \emph{Return}: Model $\Theta = \left(\{\w_i\}_{i=1}^M, \{c_j\}_{j=1}^{M!}\right)$
\caption{SGD based learning for LOMo}
\label{algoSGD}
\end{algorithmic}
\end{algorithm}

\subsection{Learning}
\label{learning}
We propose to learn the model using a stochastic gradient descent (SGD) based algorithm with
analytically calculable sub-gradients, given as
\begin{align}
\nabla_{\w_i} \L = \left\{
              \begin{array}{ll}  
                    \lambda_1 \w_i -  (1-\gamma_g)y\x_{k_i}/M & \textrm{if \ \ }y s_{\Theta}(X) < 1 \\  
                    \lambda_1 \w_i  & \textrm{otherwise.}
              \end{array} \right. 
\end{align}
The expression for gradient \wrt $\w_g$ is also similar, with $\x_{k_i}/M, (1-\gamma_g)$ replaced by
$\x_g, \gamma_g$ respectively and  $\nabla_{c_i} \L = 1$ if $ y s_{\Theta}(X) < 1 $ and $0$
otherwise.  The algorithm, summarized in \autoref{algoSGD}, randomly samples the training set and
does stochastic updates based on the current example. Due to its stochastic nature, the algorithm is
quite fast and is usable in online settings where the data is not entirely available in advance and
arrives with time. 

The scoring optimization can be solved exactly using Dynamic Programming (DP). However, in practice we found the 
DP based solver to be slow and we resorted to an approximate but much faster algorithm. In the experimental
results reported in this paper, we solve the scoring optimization with the following approximate algorithm. We obtain
the best scoring frame $\x_{k_i}$ for $\w_i$, sequentially for $i=1,\ldots,M$, and remove $\w_i$ from the model and
$\x_{k_{i-t}},\ldots,\x_{k_{i+t}}$ frames from the video; and repeat steps $M$ times so that every $\w_i$ has a
corresponding $\x_{k_i}$. $t$ is a hyperparameter to ensure temporal coverage by the model -- it stops multiple $\w_i$'s
from choosing (temporally) close frames. Using such suppression we approximately incorporate the overlap constraint in
the scoring function (\autoref{eqnScore1} \& \autoref{eqnScore2}); the hyperparameter $\beta$ is replaced by $t$. Once
the $\k = k_1,\ldots,k_M$ sub-event locations are chosen we add $c_{\sigma(\k)}$ to their average template score \ie
$\sum_{i=1}^M \w_i^\top \x_{k_i}/M$. We refer the readers to appendix for discussion regarding using DP based solution for inference.

\section{Experimental Results} 
\label{secExp}

We evaluate the proposed algorithms on two domains of
facial analysis and human actions. 
We now present the details of the datasets and the experimental settings, and then 
discuss the results. 

\subsection{Facial Analysis Datasets and Settings}
We empirically evaluate the proposed approach on four challenging, publicly
available, facial behavior datasets, of emotions, clinical pain and non-verbal
behavior, in a weakly supervised setting \ie without frame level annotations.  
The four datasets range from both posed (recorded in lab setting) to
spontaneous expressions (recorded in realistic settings). 

\subsubsection{Datasets}
\textbf{CK+ 
}
\cite{cohnkanade} is a benchmark dataset for expression recognition, with $327$ videos from $118$
participants posing for seven basic emotions -- anger, sadness, disgust, contempt, happy, surprise
and fear. We use a standard subject independent $10$ fold cross-validation and report mean of
average class accuracies over the $10$ folds. It has annotations for the apex frame and thus also
allows fully supervised training and testing.
\vspace{0.8em} \\
\textbf{Oulu-CASIA VIS 
}
\cite{zhao2011facial} is another challenging benchmark for basic emotion classification. We use the
subset of expressions that were recorded under the visible light condition. There are $480$
sequences (from $80$ subjects) and six classes (as CK+ except contempt).  It has a higher
intra-class variability as compared to CK+ due to differences among subjects. We report average
multiclass accuracy and use subject independent folds provided by the dataset creators.
\vspace{0.8em} \\
\textbf{UNBC McMaster Shoulder Pain
}
\cite{lucey2011painful} is used to evaluate clinical pain prediction. It consists of real world
videos of subjects with pain while performing guided movements of their affected and unaffected arm
in a clinical interview. The videos are rated for pain intensity ($0$ to $5$) by trained experts.
Following~\cite{Wu2015FG}, we label videos  as ``pain'' for intensity above three and ``no pain'' for
intensity zero, and discard the rest. This results in $149$ videos from $25$ subjects with $57$
positive and $92$ negative samples. Following \cite{Wu2015FG} we do a standard leave-one-subject out
cross-validation and report classification rate at ROC-EER.
\vspace{0.8em} \\
\textbf{LILiR 
}
\cite{sheerman2009feature} is a dataset of non-verbal behavior such as agreeing, thinking, in natural
social conversations. It contains $527$ videos of $8$ subjects involved in dyadic conversations. The
videos are annotated for $4$ displayed non-verbal behavior signals- agreeing, questioning, thinking
and understanding, by multiple annotators. We generate positive and negative examples by
thresholding the scores with a lower and higher value and discarding those in between. We then
generate ten folds at random and report average Area under ROC -- we will make our cross-validation
folds public. This differs from Sheerman \etal \cite{sheerman2009feature}, who used a very small
subset of only $50$ video samples that were annotated with the highest and the lowest scores. 

\subsubsection{Features}
We compute four types of facial descriptors. We extract
$49$ facial landmark points and head-pose information using supervised gradient
descent
\cite{xiong2013supervised} and use them for aligning faces. The first set of descriptors are 
SIFT-based features, which we compute by extracting SIFT features around facial landmarks and
thereafter concatenating them \cite{xiong2013supervised, de2011facial}. We align the faces into
$128 \times 128$ pixel and extract SIFT features (using open source \texttt{vlfeat}
library
\cite{vedaldi08vlfeat}
) in a fixed window of size $12$ pixels. The SIFT features are normalized to
unit $\ell_2$ norm. We chose location of $16$ landmark points around eyes ($4$), brows ($4$), nose ($2$)
and mouth ($6$) for extracting the features. Since SIFT features are known to contain redundant
information \cite{ke2004pca}, we use Principal Component Analysis to
reduce their dimensionality to $24$. To each of these frame-level features, we add coarse temporal
information by appending the descriptors from next $5$ consecutive frames, leading to a
dimensionality of $1920$. The second features that we use are geometric features
\cite{zhang1998comparison,de2011facial}, that are known to contain shape or location information of
permanent facial features (\eg eyes, nose). We extract them from each frame by subtracting $x$ and
$y$ coordinates of the landmark points of that frame from the first frame (assumed to be neutral)
of the video and concatenating them into a single vector ($98$ dimensions). We also compute LBP
features 
(with radius $1$ and neighbourhood
$8$) that represent texture information in an image as a histogram. We add spatial information to
the LBP features by dividing the aligned faces into a $9 \times 9$ regular grid and concatenating
the histograms ($4779$ dimensions) \cite{sikka2012,lazebnik2006beyond}. We also consider Convolution
Neural Network (CNN) features by using publicly available models of Parkhi \etal \cite{Parkhi15}
that was trained on a large dataset for face recognition. We use the network output from the
last fully connected layer. However, we found that these performed lower than other features \eg on
Oulu and CK+ datasets they performed about $10\%$ absolute lower than LBP features. We suspect
that they are not adapted to tasks other than identity discrimination and did not use them further.

\subsubsection{Baselines}

For our experiments on human facial behavior analysis we report results with $4$ baseline
approaches. For first two baselines we use average (or mean) (MnP) and max temporal pooling (MxP)
\cite{sikkaexemplar} over per-frame facial features along with SVM. Temporal pooling is often used
along with spatio-temporal features such as Bag of Words
\cite{laptevCVPR2008,sikka2014classification}, LBP \cite{zhao2007dynamic} in video event
classification, as it yields vectorial representation for each video by summarizing variable length
frame features. We select Multiple Instance Learning based on latent SVM \cite{andrews2002support}
as the third baseline algorithm. We also compute the performance of the fully supervised (FS) algorithms
for cases with known location of the frame that contains the apex expression. For making a fair
comparison, we use the same implementation for SVM, MIL and LOMo. 

\subsubsection{Parameters} We fix $M=1$ and $c_{\sigma}=0$ in the current implementation, for obtaining SVM baseline
results with a single vector input (mean and max pooling), and report best results across both learning rate and number of iterations. For both
MIL ($M=1$) and LOMo, which take a sequence of vectors as input, we set the learning rate to $\eta=0.05$ and for MIL we
set $c_\sigma=0$. We fix the regularization parameters $\lambda_1=10^{-5}$ and $\lambda_2=0$  for all experiments, based
on initial validation experiments. We do multiclass classification using one-vs-all strategy. For ensuring temporal
coverage (see \autoref{learning}), we set the search space for finding the next sub-event to exclude $t=5$ and $50$
neighboring frames from the previously detected sub-events' locations for datasets with fewer frames per video (\ie CK+,
Oulu-CASIA VIS and LILiR datasets) and UNBC McMaster dataset, respectively. We did not do (dataset specific)
cross-validation for these hyperparameters as we found the results to be stable across different choice of
hyperparameters owing to similar domain of the datasets. We set $M=3$ for all datasets except CK+ where we set
$M=2$ since it consists of posed expressions containing only onset and apex phase spanning a few
frames. For our final implementation, we combine LOMo models, learned independently on different
features, using late fusion \ie we averaged the prediction scores with equal weights.  

\subsection{Human Actions Datasets and Settings }
We also evaluate our approach on $3$ challenging publicly available human action classification
datasets. Their videos has only video-level labels and cover wide range of activities such as
sports, grooming, human interactions, that showed wide variability in appearance, temporal
structure, duration, viewpoints \etc 

\vspace{0.8em} 
\textbf{Olympic Sports}
\cite{niebles2010modeling} is a dataset of sports activities 
\eg snatch, clean and jerk, high jump \etc It contains $783$ video samples from $16$
sports classes with most videos collected from YouTube. Most of these the activity classes are
complex activities \cite{niebles2010modeling, turaga2008machine} in that they are composed of
simpler actions, \eg long jump activity is composed of standing, running and jumping. In addition to
simple limb movements this dataset also involves interactions between humans and objects, \eg javelin
in javelin throw class. We use the train and test splits provided by the authors and report Mean
Average Precision (mAP) across all classes.  
\vspace{0.8em} \\ 
\textbf{High Five
}
\cite{patron2010high} dataset consists of videos of interactions between humans which were
collected from TV shows. The datasets contains $200$ clips from $4$ classes, \ie hug, kiss, handshake,
highfive and $100$ clips from a negative class. This dataset was introduced to study two person
interactions as opposed to single person interaction. We use the train and test folds provided by
the authors and report mAP across two fixed cross-validation folds.  

\vspace{0.8em} 
\textbf{HMDB}
\cite{kuehneICCV2011} dataset contains wide range of human actions including facial actions
(\eg smile, laugh), facial actions with object manipulation (\eg smoke, drink), general body
movements (\eg cartwheel, clap hands, climb, jump), body movements with object interactions (\eg
catch, brush hair, shoot bow), and body movements for human interactions (\eg fencing, hug, kiss).
It has around $6800$ clips from $51$ classes, collected from YouTube, Prelinger
archive \etc The dataset is very challenging; it has videos with significant camera motion
($59.9\%$), poor quality (only $17\%$ high quality clips), and non frontal camera viewpoints ($40\%$
clips have frontal viewpoint). We report both mAP and mean multiclass accuracy across $3$ cross-validation
folds provided by the authors. 

\subsubsection{Features}

We use both Improved Dense Trajectory (iDT) \cite{wangIJCV2015} and CNN based features
\cite{SimonyanICLR2015, he2015deep, tran2015learning} for our experiments on classifying human
actions. We extract the iDT features using the tool provided by the authors and also use the available
human bounding boxes for HMDB and Olympic dataset \cite{wangIJCV2015}.  For highFive dataset we use
the bounding boxes provided by the dataset creators \cite{patron2010high}. We extract iDT features
with trajectory length of $15$ for HMDB and highFive dataset, while for Olympic dataset we use
trajectories of length $5$ as used in \cite{gaidonIJCV2014}, as trajectories for fast moving motions
(especially sports) are usually unable to describe salient objects for more than a few frames.  The
trajectories are described using four features, \ie HOG, HOF and  MBH along X-axis and Y-axis. We use PCA
to reduce the dimensionality of these features by half and encoded them separately using Fisher
Vector (FV) encoding with dictionaries of $64$ elements.  We then perform power normalization
($\alpha = 1/2$) and $L2$ normalization for the FVs \cite{perronninECCV2010} and concatenate the
FVs for different low-level features as the final representation (dimensionality is $25344$). We
refer to these features as iDT-64-FV denoting the different components used.  For both clustering
via Gaussian Mixture Modeling and extracting FV we use the \texttt{vlfeat} library
\cite{vedaldi08vlfeat}.
We construct segment-level local features by using a temporal window of size of $W$ around each
frame and then compute the above FVs on features lying in the temporal window. This averaging
results in adding
temporal context to each individual feature. The length of $W$ determines the duration of local
sub-events and we set it based on the average duration of each clip and an estimate of the duration of
corresponding local temporal events in each dataset.  We set $W=11$ for Olympic dataset, $W=5$ for
highFive dataset and $W=3$ for HMDB dataset. The Olympic dataset in general include longer videos
with average duration of $\sim 230$ frames compared to $\sim 90$ in HighFive and HMDB. Also, the
local temporal events in general are longer in Olympic dataset compared to HighFive and HMDB.  We
use the above procedure to obtain global temporal features for the Adaptive LOMo algorithm by using
a temporal window that includes the entire duration of a videos.  We extract CNN features from
$16$ layer VGG network \cite{SimonyanICLR2015} and $152$ layer ResNet \cite{he2015deep}, both trained for
image classification on ImageNet dataset.  For each frame, we resize it such that the smallest
dimension is of size $256$ and then crop a central $224 \times 224$ region.  We then compute outputs
from fc-6 layer for VGG network (dimensionality is $4096$) and pool-5 layer of ResNet
(dimensionality is $2048$). We also extract spatio-temporal C3D features as described in
\cite{tran2015learning} by using a network that was pretrained on Sports dataset. 
We use Caffe implementation provided by Tran \etal and extract features
from fc-6 later over a $16$
frame window. We use the same procedure as outlined for FVs for obtaining local and
global temporal features except we use mean temporal pooling and only perform $L2$ normalization in
the end.

%
\subsubsection{Baselines} 

We report results with five baselines for evaluation of Adaptive LOMo (A-LOMo) on human activity classification. The
first baseline is global temporal pooling (GTP) which is obtained by using global temporal features. The second baseline
is LOMo that was obtained with local temporal features only. We also compare with MIL algorithm as done for facial
analysis. The fourth baseline is with LOMo with the ordinal component removed and the final baseline is 
Adaptive LOMo with $M=1$ (MIL+GTP), which is essentially combining MIL with GTP.
%
\subsubsection{Parameters} Since the human activity datasets varied in the number and type of classes, viewpoints,
camera motion, duration of clips and number of samples, we opt for using $5$ fold cross validation for setting $\lambda$
parameter. Further, since different actions have events that span different durations, we also set the temporal coverage
parameter  $t$ by the same cross-validation. We sweep $t$ from $5$ to $50$ and during inference we set $t$ to minimum of
the cross-validated value and number of frames divided by the number of events. This is done to handle clips whose
duration are smaller than the $t$ parameters. In order to reduce the number of cross validation steps we first set
$\lambda$ and $t$ using cross-validation for LOMo and then fix these values for cross-validating the $\gamma_g$
parameter in Adaptive LOMo. We observed during our experiments that the results were not very sensitive to these
parameters. The second regularization parameter $\lambda_2$ is set to $10^{-3}$ in our experiments. 
We set $M=3$ across all experiments based on our initial validation experiments
where we observed that in general the results were stable across different
classes for $M=3$ and then begin to drop. 
The GTP based baseline is obtained by setting $\gamma_g = 1$ in Adaptive LOMo algorithm,
while the LOMo baseline is obtained by setting $\gamma_g = 0$. We report the results for MIL baseline by setting $M=1$
and $c_\sigma=0$. For our final implementation we combine Adaptive LOMo models, learned independently on different
features, by weighted averaging of z-score normalized prediction scores (late fusion).  
We obtained the weights by doing a coarse grid search over the set $[0, 0.5, 1]$. The weights
obtained for the three datasets were $0.5$ for ResNet based features, $0.5$ for C3D (except for $0$
in HighFive dataset), $0.5$ for Objects (used in HMDB dataset), and $1$ for for iDT.

\subsection{Quantitative Results}
\subsubsection{Facial Behavior Analysis}
\label{expFaceQuant}
The performances of the proposed approach, along with those of the baseline methods, are shown in
\autoref{tab_lomo_1}. In this comparison, we use SIFT-based facial features for all datasets.
Since head nod information is important for identifying non-verbal behavior such as agreeing, we
also append head-pose information (yaw, pitch and roll) to the SIFT-based features for the LILiR
dataset. 

We see performance improvements with proposed LOMo, in comparison to baseline methods, on $6$ out of
$7$ prediction tasks. In comparison to MIL, we observe that LOMo outperforms the former method on
all tasks. The improvements are $1.2\%, 4.2\%$ and $1.1\%$ absolute, on CK+, Oulu-CASIA VIS and UNBC
McMaster datasets, respectively. This improvement can be explained by the modeling advantages of
LOMo, where it not only discovers multiple discriminative sub-events but also learns their ordinal
arrangement. For the LILiR dataset, we see improvements in particular on the `Questioning' ($5.9\%$
absolute) and `Agreeing' ($1.7\%$ absolute), where temporal information is useful for recognition.
In comparison to temporal pooling based approaches, LOMo outperforms both variants of temporal
pooling- MnP and MxP,  on $6$ out of $7$ tasks. This is not surprising since temporal pooling
operations are known to add noise to discriminative segments of a video by adding information from
non-informative segments \cite{sikkaexemplar}. Moreover, they discard any temporal ordering, which
is often important for analyzing facial activity \cite{sikka2014classification}.
\begin{table}[t]
\centering
\newcolumntype{C}{>{\centering\arraybackslash}p{2.5em}}
\scalebox{1.0}{
\begin{tabular}{|r|r|C|C|C|C|C|}
\hline
Dataset         &Task         & FS         & MnP        & MxP & MIL     & LOMo \\ \hline \hline
CK+             &Emot.& \textbf{$91.9$}    & $86.0$           & $87.5$      &$90.8$   &$\textbf{92.0}$ \\ \hline 
O-C VIS  &Emot.& \textbf{$75.0$}    & $68.3$           & $69.0$      &$69.8$   &$\textbf{74.0}$ \\ \hline 
BC-McM &Pain                  & $-$  & $67.4$           & $81.5$      &$85.9$   &$\textbf{87.0}$ \\ \hline 
\multirow{4}{*}{LILiR}  &Agree     & $-$  & $84.7$  & $\textbf{85.5}$      &$77.7$   &$79.4$ \\ \cline{2-7}
                &Ques.& $-$  & $86.2$           & $84.3$      &$80.7$   &$\textbf{86.6}$ \\ \cline{2-7}
                &Think.& $-$  & $93.6$           & $88.9$      &$93.8$   &$\textbf{94.8}$ \\  \cline{2-7}
                &Under.& $-$  & $79.4$           & $79.2$      &$78.9$   &$\textbf{80.3}$ \\ \hline
        \end{tabular}
}
\caption{Comparison of LOMo with baselines on the facial analysis datasets using SIFT based features (see \autoref{expFaceQuant}).}
\label{tab_lomo_1}
\vspace{-1em}
\end{table}

On both facial expression tasks, \ie emotion (CK+ and Oulu-CASIA VIS) and pain prediction (UNBC
McMaster), methods can be arranged in increasing order of performance as MnP, MxP, MIL, LOMo. A
similar trend between temporal pooling and weakly supervised methods has also been reported by
previous studies on video classification \cite{sikka2014classification,gaidonPAMI2013}.  We again
stress that LOMo performs better than the existing weakly supervised methods, which are the
preferred choice for these tasks. In particular, we observe the difference to be higher between
temporal pooling and weakly supervised methods on the UNBC McMaster dataset, \ie $67.4\%$ for
MnP, $81.5\%$ for MxP, $85.9\%$ for MIL and $87.0\%$ for LOMo. This is because the
subjects exhibit both head movements and non-verbal behavior unrelated to pain, and thus focusing on
the discriminative segment, \cf using a global description, leads to performance gain. However, we
do not notice a similar trend on the LILiR dataset -- the differences are smaller or reversed, \eg
for `Understanding' mean-pooling is marginally better than MIL ($79.4\%$ \vs $78.9\%$), while LOMo
is modestly better than both ($80.3\%$).  This could be because most conversation videos are
pre-segmented and predicting non-verbal behavior relying on a single prototypical segment might be
difficult, \eg `Understanding' includes both upward and downward head nod, which cannot be captured
well by detecting a single event. In such cases we see LOMo beats MIL by temporal modeling of
multiple events.   
We also performed t-test between the results on MIL and LOMo and observed the p-values were low
($\leq$ $5\%$) for Oulu and LiLiR- questioning task.  The p-values were moderately low ($\leq$
$15\%$) for CK+ and LiLiR-think. We account higher p-values in other cases to a small number of
samples in most datasets \eg $149$ in UBC-McMaster. We also observed the results of LOMo to be
higher or equal to MIL in most test folds on all the datasets \eg $21/25$ for UNBC McMaster.

\subsubsection{Human Action Classification}
\label{expActQuant}

The comparison of Adaptive LOMo with several baseline methods on $3$ datasets and two different features is shown in
\autoref{tab_lomo_2}.  When using iDT features we see that Adaptive LOMo outperforms the baseline methods on all the
datasets except HMDB. The mAP score for Adaptive LOMo is $88.3\%$ on the Olympic dataset as compared to $84.6\%$ with
LOMo, $79.6\%$ with MIL, $85.4\%$ with GTP, $75.8$ with LOMo without ordinal component, and $86.5$ with MIL+GTP. We also
see that on HMDB dataset, both Adaptive LOMo ($52.2\%$) and LOMo ($48.2\%$) outperforms MIL ($44.8\%$) by a significant
margin showing the benefits of using the proposed local structure instead of a single discriminative segment. These
results clearly demonstrate that, by learning to adapt to different classes, Adaptive LOMo combines the strength of both local
and global temporal structure and results in performance improvements. The baseline with MIL+GTP is
also outperformed by Adaptive LOMo in majority of the cases, further demonstrating that the local
ordinal structure is important in the combination of local and global information as well.

\begin{table}[t]
\centering
\newcolumntype{C}{>{\centering\arraybackslash}p{2.5em}}
\scalebox{1.0}{
        \begin{tabular}{|r|r|C|C|C|C|C|C|}
        \hline
	Dataset &Feats&GTP & MIL & LOMo & LOMo (ord=0)  & MIL+ GTP & A-LOMo\\
        \hline \hline
	\multirow{2}{*}{Olympic}    &iDT-64-FV   &$85.4$  &$79.6$   &$84.6$  & $75.8$ & $86.5$  &$\mathbf{88.3}$ \\
                                    &CNN        &$69.8$      &$63.7$ &$69.8$    & $70.0$  & $68.7$    &$\mathbf{71.3}$ \\ \hline
        \multirow{2}{*}{HighFive}   &iDT-64-FV   &$61.1$      &$59.0$    &$58.1$     & $54.4$  & $62.1$      &$\mathbf{64.8}$ \\
                                    &CNN        &$31.8$      &$\mathbf{35.3}$        &$35.0$   & $34.5$ & $34.8$   &$35.0$ \\ \hline
	\multirow{2}{*}{HMDB}       &iDT-64-FV    &$50.5$     &$44.8$     &$48.2$    & $45.1$ & $\mathbf{54.1}$       &$52.2$ \\
				                  &CNN        &$34.4$      &$34.4$    &$34.5$    & $33.1$ & $\mathbf{35.9}$    &$35.4$ \\ \hline
\end{tabular}
}
\caption{Comparison of LOMo with baselines on three human activity datasets.
The performance reported is mean average
precision (mAP) (see \autoref{expActQuant}).}
\vspace{-1em}
\label{tab_lomo_2}
\end{table}

In terms of features, the improvements relative to GTP are higher for iDT features as compared to CNN
features. For example, the relative improvement between Adaptive LOMo and global temporal pooling is $3.4\%$
for iDT features and $2.1\%$ for CNN features on the Olympic dataset. This trend for performance
improvement is similar for the HMDB dataset ($3.4\%$ for iDT \vs $2.9\%$ of CNN). We explain this
observation by the presence of motion information in the iDT features, that results in more
meaningful local temporal segments and benefits our algorithm. This is particularly true for
classes where motion cues seem to be more important for discrimination compared to appearance cues
\cite{kuehneICCV2011}. For example, we observed the relative improvement between Adaptive LOMo and GTP
across iDT and CNN features to be high for classes such as ``chew'' ($4.7\%$
for iDT \vs none for CNN), ``shootbal'' ($14.9\%$ for iDT \vs none for CNN), ``handstand'
($26\%$ for iDT vs $14.4\%$ for CNN), ``highjump'' ($43.4\%$ for iDT vs $13.6\%$ for CNN).

\begin{figure}[t]
        \centering
        \includegraphics[width=0.9\columnwidth,trim=0 0 0 
        0, clip]{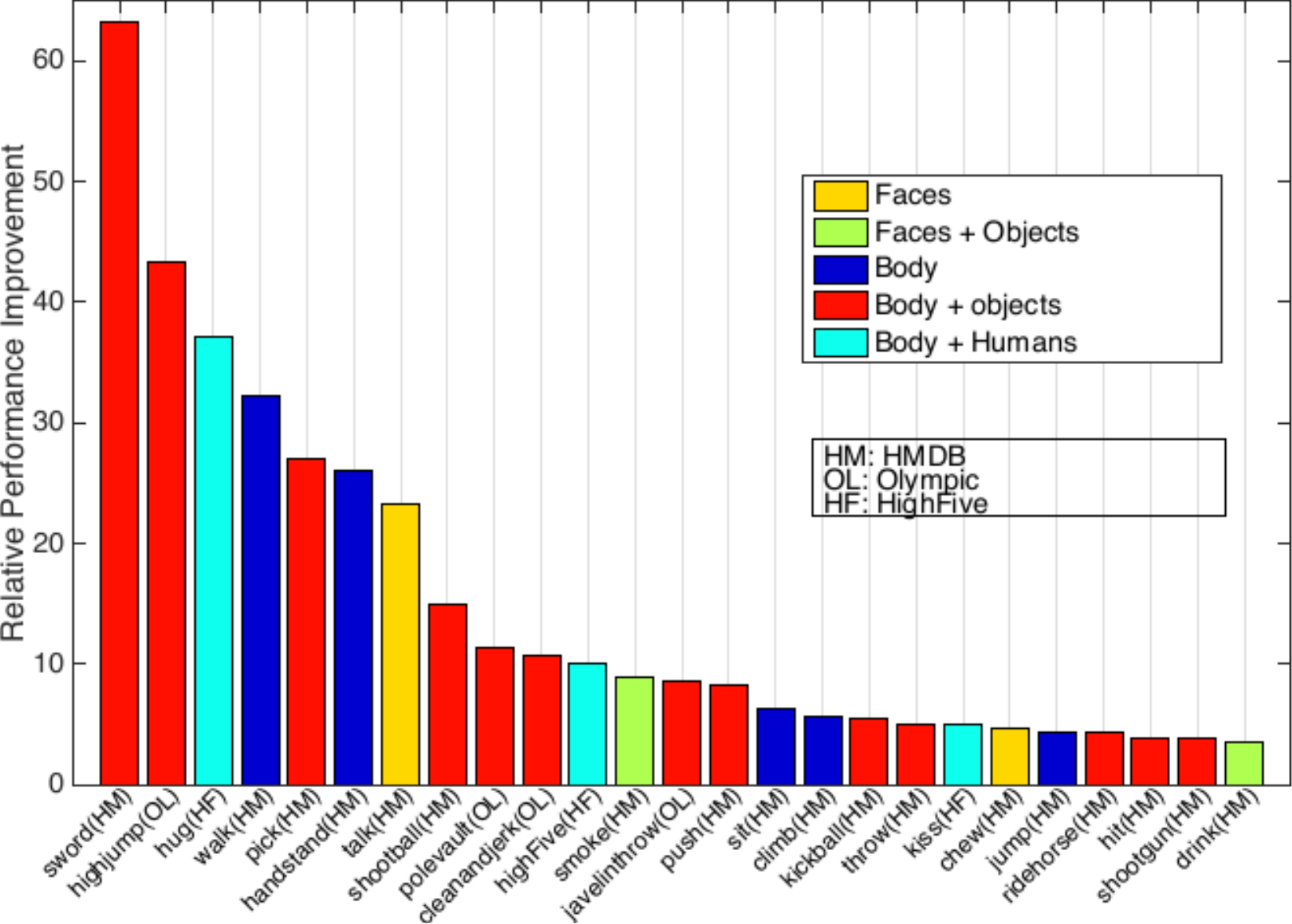}
        \caption{The relative improvements for different action classes for Adaptive LOMo \vs global
        temporal pooling, both with iDT features. Color codes indicate the type of
        action and the respective datasets are in brackets.}        
        \label{plot_improve}
        \vspace{-.5em}
\end{figure}

The HMDB dataset was collected by asking students to annotate parts of a clip that represented a
single non-ambiguous human action \cite{kuehneICCV2011}. Thus it is mostly composed of pre-segmented
clips where for certain classes the appearance information may not vary much in comparison to the
motion information.  For example, in the class ``shootball" appearance features such as CNN may always
encode a person, basketball net and a basketball, and this representation may not vary much across
frames. On the other hand, motion information can effectively encode the movement of the person and
the ball. This information can then effectively represent sub-events, such as standing, jumping and
shooting the ball, that describe this class. This is the reason for our method, that combines GTP
with local sub-events, to gain higher performance improvements over GTP while using iDT
features. On HMDB dataset we observed that Adaptive LOMo outperforms GTP on $34$ out of $51$ classes
when using iDT features. On the other hand, with CNN features it outperformed only in $12$ classes. 
We saw a similar trend for HighFive dataset where Adaptive LOMo outperformed GTP on
all classes with iDT features, and on $3$ out of $4$ classes with CNN features. On the Olympic dataset,
Adaptive LOMo trained with iDT features showed improvement on $12$ of $16$ classes, while with CNN features
it showed an improvement on $5$ out of $16$ classes. We also see that LOMo outperforms its
counterpart without ordinal information, while using iDT features, on all datasets. These trends
clearly highlight that motion information is critical for learning local sub-events and their
ordinal relationship is important as well. For the remaining classes we observed that the results
were either similar to GTP or lower when cross-validation was unable to estimate the
correct weighting factor.  

We have also shown some classes with the best relative improvement between Adaptive LOMo and GTP 
in \autoref{plot_improve}. These classes show improvement over baseline due to the
addition of LOMo that models each video as a collection of local sub-events with loose ordering.
We observe that the classes with most improvements can either be decomposed into short temporal
segments, \eg ``hug'', ``highjump'', ``pick'', or involve repetitions of simpler motion segments, \eg
``talk'', ``walk'', ``climb''. We conclude from from this observation that our method generalizes to
not only activities composed of simple motion segments but also to categories that involve repetitions of
these simpler actions. Motion information is crucial for discriminating between
fine-grained classes such as sword drawing \vs sword fighting, and our method is able to learn and
yield performance improvement from this information. Other example of such classes are ``cleanandjerk''
and ``poleactivity'' that seem similar to the class ``snatch'' in appearance but differ in composition of the
temporal segments. We observed during our experiments that for some classes Adaptive LOMo selected
either mostly local or mostly GTP. Example of classes where mostly one of the
two components was selected are: (i) local ($\gamma_g \leq 0.2$) -- ``cleanandjerk", ``hug", ``handstand", ``talk",
``highjump'', ``shootball'', ``pullup'', ``pushup'' and (ii) global ($\gamma_g \geq 0.8$) -- ``turn'', ``drawsword'', ``kiss'',
``shakehands'', ``catch'', ``kickball'', ``eat'', ``wave''. Notice, that in the later classes the action is expected to
be spread out in the whole duration of the video. We believe that being able to adapt the classifier
to different classes not only results in improvement, as seen with Adaptive LOMo, but also explain
the underlying temporal structure. 

\subsection{Qualitative Results}
\label{secQualRes}

We now give some qualitative results in this section by showing detections on some test samples.
The frames are overlaid with detections from the discovered sub-events with Event 1 in red color,
Event 2 in green color and Event 3 in blue color. Since the LOMo model starts with random
initialization the events can vary across classes in terms of both the most probable ordering and
their semantics.  

\begin{figure}[t]
        \centering 
        \includegraphics[width=\columnwidth,trim=180 385 1300 200,clip]{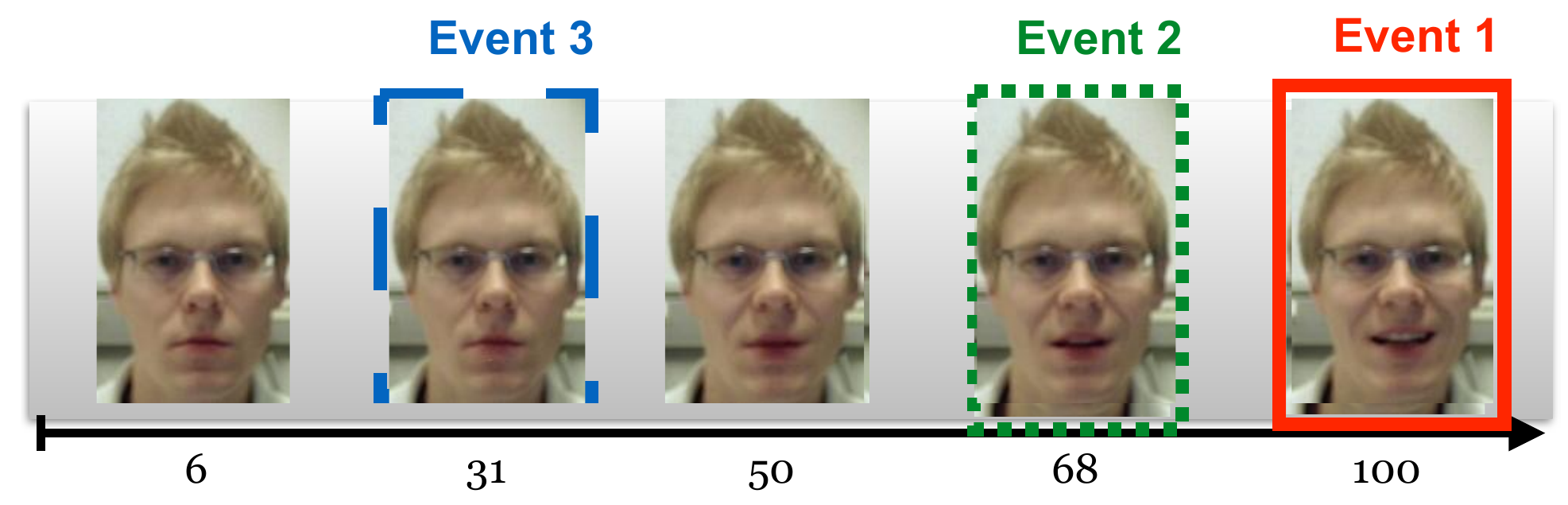} \\
        \includegraphics[width=\columnwidth,trim=180 385 1300 200,clip]{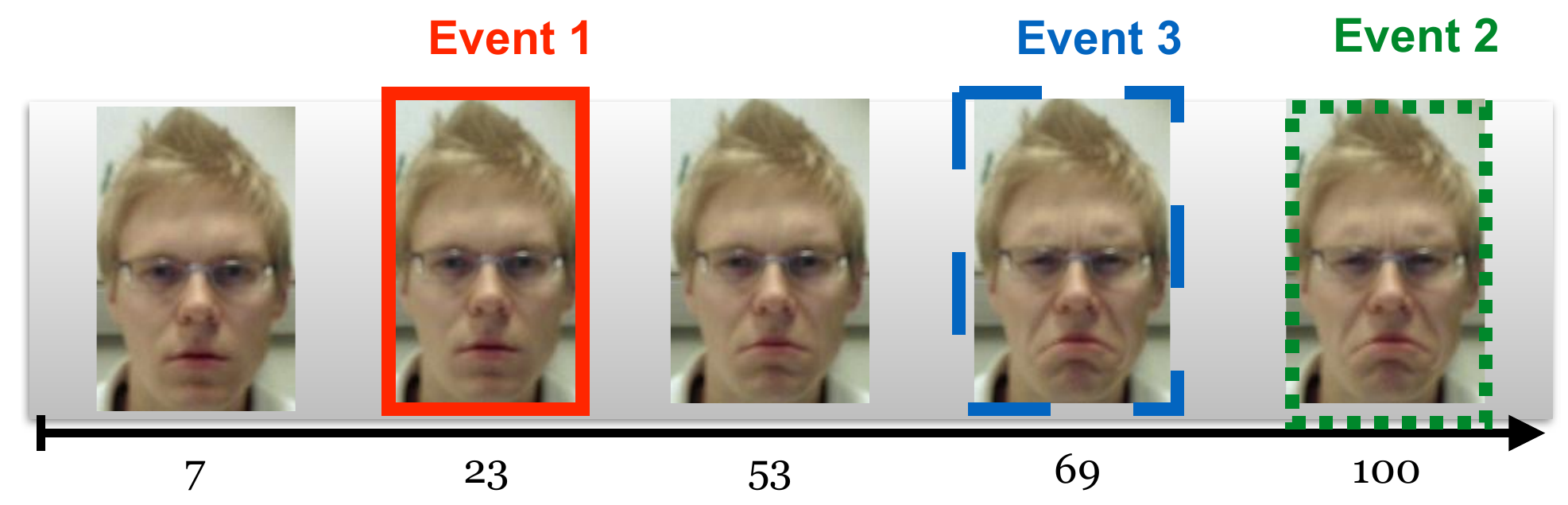} \\
\vspace{-.5em}
\caption{Detections made by LOMo trained ($M=3$) for classifying `happy' expression on two
	expression sequences from Oulu-CASIA VIS dataset (ground-truth for top
	is `happy' and bottom is `sad'). The number below the timeline shows
	the relative location
(in percentile of total number of frames).} 
\vspace{-1em}
\label{figoulu_hap}
\end{figure}

\begin{figure*}
        \includegraphics[width=\textwidth,trim=180 385 380 210,clip]{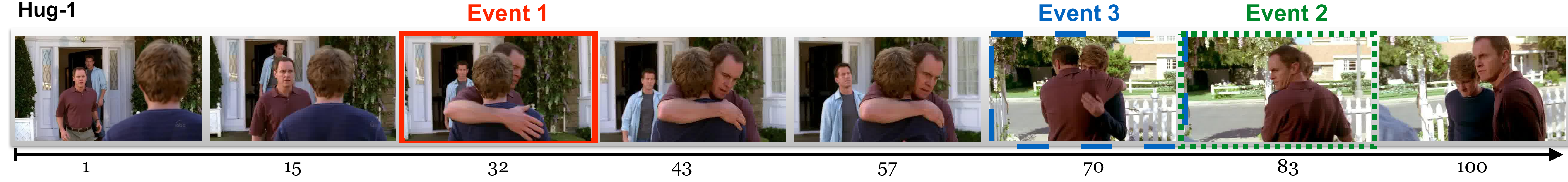}       

        \vspace{-0.05em}

        \includegraphics[width=\textwidth,trim=180 385 380 200,clip]{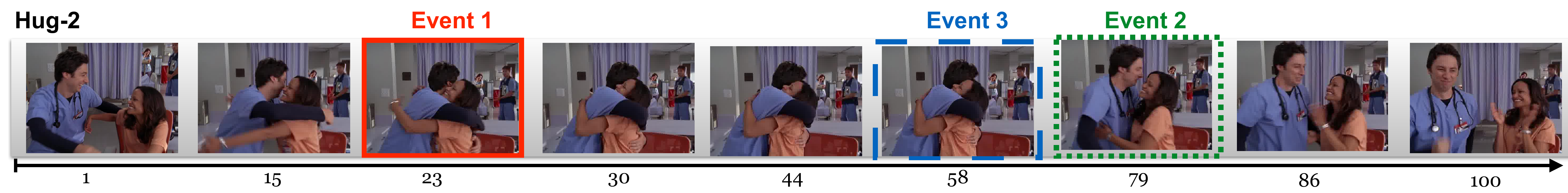}  

        \includegraphics[width=\textwidth,trim=180 385 380 180,clip]{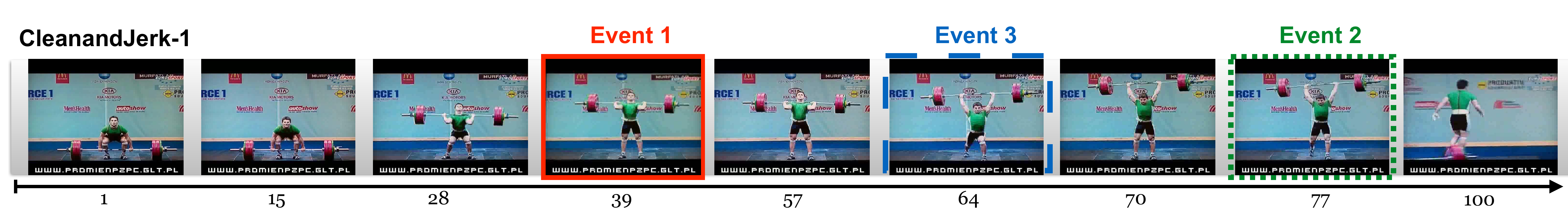}       

        \vspace{-0.16em}

        \includegraphics[width=\textwidth,trim=180 385 380 190,clip]{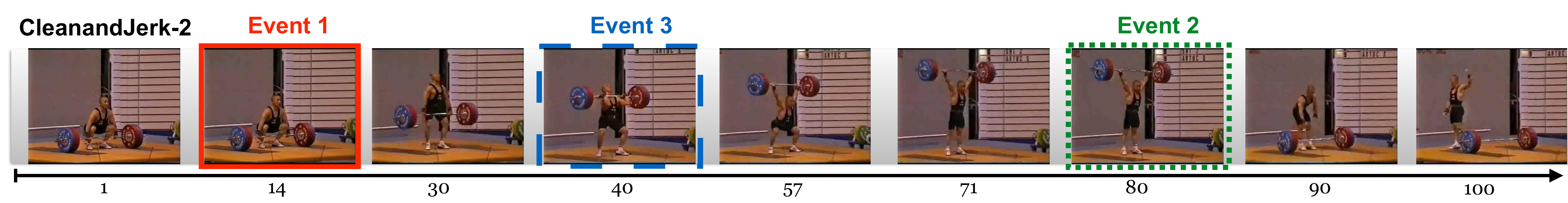}       

        \includegraphics[width=\textwidth,trim=180 385 380 180,clip]{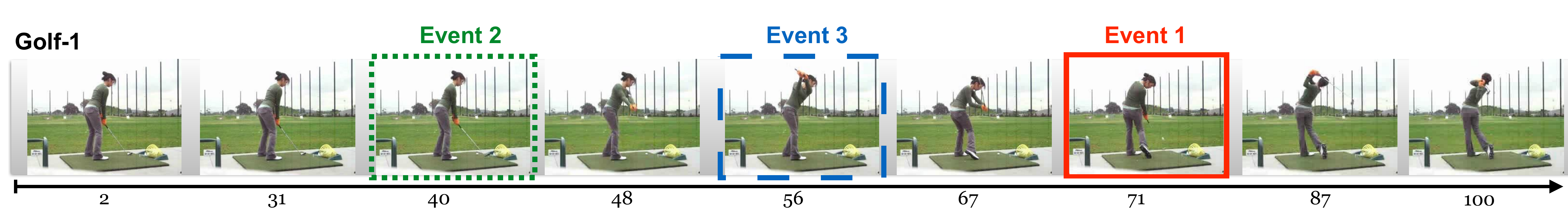}       

        \vspace{-0.18em}

        \includegraphics[width=\textwidth,trim=180 385 380 190,clip]{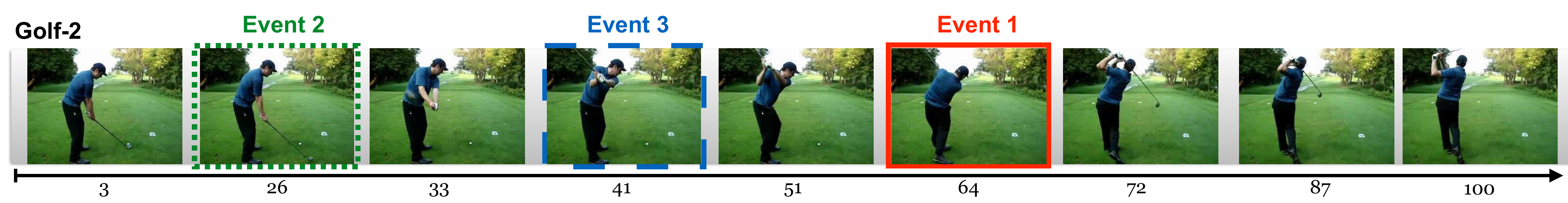}       

        \includegraphics[width=\textwidth,trim=180 385 380 180,clip]{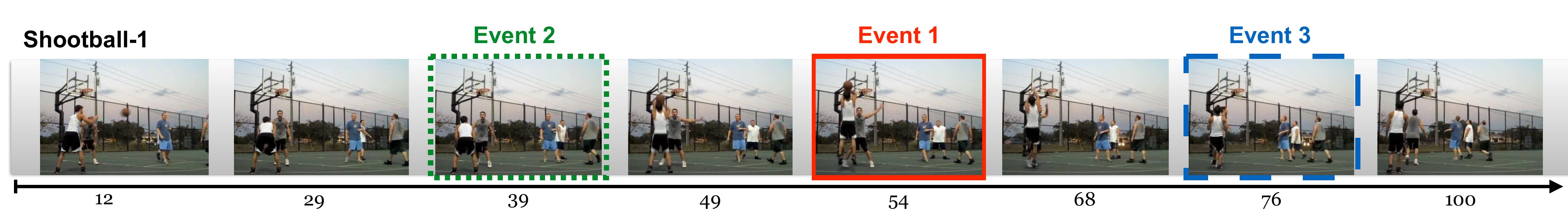}       

        \vspace{-0.18em}

        \includegraphics[width=\textwidth,trim=180 385 380 190,clip]{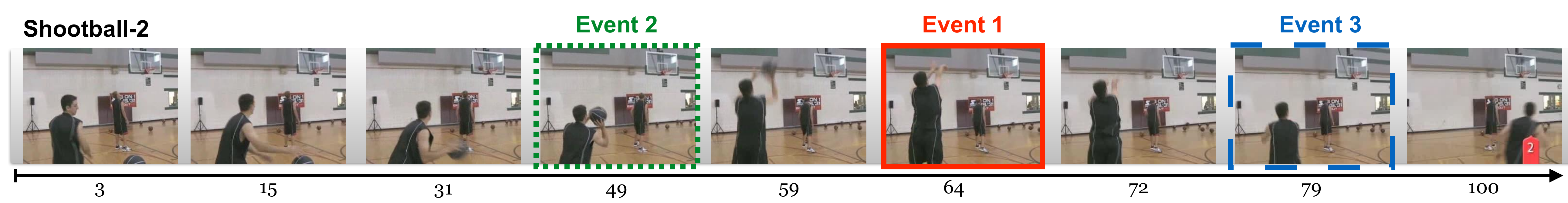}       
        \vspace{-2em}
        \caption{Detection of multiple discriminative sub-events, discovered by LOMo, on the three
                different human action analysis datasets. The number below the timeline shows the relative
        location (in percentile of total number of frames).}
        \vspace{-1em}
        \label{human_action_qual}
\end{figure*}

\subsubsection{Facial Behavior Analysis}

\autoref{figoulu_hap} shows the detections of our approach, with model trained for ``happy''
expression, on two sequences from the Oulu-CASIA VIS dataset. The model is trained with three
sub-events. As seen in \autoref{figoulu_hap}, the three events seem to correspond to the expected
semantic events, \ie neutral, low-intensity (onset) and apex, in that order, for the positive
example (left), while for the negative example (right) the events are incorrectly detected and are
in the wrong order as well. To further illustrate the importance of learning the ordering, we
observed that the ordering cost learned by the model for the ordering $(1,3,2)$ was $-0.6$ which
is much lower than $0.9$ for the correct order of $(3,2,1)$, as observed in the positive happy
example.  This result highlights the modeling strength of LOMo, where it learns both multiple
sub-events and a prior on their temporal order. 


\subsubsection{Human Action Classification}

We have shown detections on test samples in \autoref{human_action_qual} from class ``hug'' in HighFive
dataset, class ``cleanandjerk'' in Olympic dataset and classes ``golf'' and ``shootball'' from HMDB dataset. We
see the consistency between sub-events detected for these classes. For example for ``golf'' 
class the three events seem to focus on backward motion with club, beginning of forward motion and
hitting the ball with the club respectively. We also see that since our model is learning
to focus only on (discriminative) frames that correspond to underlying activity, it can effectively filter out noisy or
irrelevant frames. For example, it is filtering out the last few frames in example ``hug-1'' and first
few frames in example ``shootball-1'' and ``Shootball-2'' where the person seems to be receiving the ball
instead of shooting it. Similar to human facial behavior, we found that the ordering cost learns to
penalize some orderings more than others. For example, in the case of class ''cleanandjerk'' the model
allows for swaps between Event 2 and Event 3, which seems to correspond to lifting up motion, but
penalizes if Event 3 or Event 2 comes before Event 1 (where the person is beginning to lift).    

Thus, we conclude that qualitatively our model supports our intuition, that not only the correct
sub-events but their correct temporal order is critical for high performance in such tasks.

\begin{table*}[t]
\centering
\newcolumntype{C}{>{\centering\arraybackslash}p{3em}}
\label{tab_lomo_21}
 \scalebox{1.0}{
\begin{tabular}{|r|C|}
\multicolumn{2}{c}{CK+ dataset \cite{cohnkanade} \vspace{0.2em}}     \\ 
\hline      
3DSIFT \cite{scovanner2007}                                   &$81.4$ \\ \hline 
LBPTOP \cite{zhao2007dynamic}                            & $89.0$ \\ \hline        
HOG3D  \cite{klaser2008spatio}                              & $91.4$ \\ \hline               
Ex-HMMs \cite{sikkaexemplar}                      & $93.9$         \\ \hline  
STM-ExpLet \cite{liu2014learning}                          & $94.2$         \\ \hline           
LOMo (proposed)& $\mathbf{95.1}$        \\ \hline      
\end{tabular}
}
\quad \quad \quad
\label{tab_lomo_22}
 \scalebox{1.0}{
\begin{tabular}{|r|C|}
\multicolumn{2}{c}{Oulu-CASIA VIS dataset \cite{zhao2011facial} \vspace{0.2em}}     \\ 
\hline
HOG3D  \cite{klaser2008spatio}                              & $70.6$ \\ \hline    
LBPTOP \cite{zhao2007dynamic}                            & $72.1$ \\ \hline       
STM-ExpLet \cite{liu2014learning}                          & $74.6$         \\ \hline    
Atlases \cite{guo2012dynamic}                               & $75.5$ \\ \hline   
Ex-HMMs \cite{sikkaexemplar}                      & $75.6$         \\ \hline      
LOMo (proposed)                                            & $\mathbf{82.1}$        \\ \hline      
\end{tabular}
}
\quad \quad \quad
\label{tab_lomo_23}
 \scalebox{1.0}{
\begin{tabular}{|r|C|}
\multicolumn{2}{c}{UNBC McMaster dataset \cite{lucey2011painful} \vspace{0.2em}}     \\ 
\hline
Ashraf et al.  \cite{lucey2008}                                       & $68.3$       \\ \hline 
Lucey et al.  \cite{lucey2008}                                        & $81.0$     \\ \hline                                         
MS-MIL \cite{sikka2014classification}                        & $83.7$    \\ \hline 
MIL-HMM \cite{Wu2015FG}                                    & $85.2$     \\ \hline 
RMC-MIL \cite{ruiz2014regularized}                         & $85.7$  \\ \hline 
LOMo (proposed) & $\mathbf{87.0}$        \\ \hline   
\end{tabular}
}
\vspace{-0.9em}
\caption{Comparison of the proposed approach with several state-of-the-art methods on three
facial analysis datasets. }
\vspace{-0.5em}
\label{tab_lomo_24}
\end{table*}

\begin{table*}[t]
\centering
\newcolumntype{C}{>{\centering\arraybackslash}p{3em}}
\label{tab_lomo_31}
 \scalebox{1.0}{
\begin{tabular}{|r|C|}
\multicolumn{2}{c}{HighFive dataset \cite{patron2010high} \vspace{0.2em}}     \\ 
\hline      
Hoai \etal \cite{hoai2014talking}                                         & $56.3$             \\ \hline      
Gaidon \etal \cite{gaidonIJCV2014}                                    & $62.4$             \\ \hline      
Ma \etal \cite{ma2015space}                                    & $64.4$             \\ \hline      
Wang \etal \cite{wangIJCV2015}                                          & $69.4$        \\ \hline      
Hoai \etal \cite{hoai2014improving}                            & $\mathbf{71.1}$        \\ \hline      
A-LOMo (proposed)                                                              & $\mathbf{70.0}$             \\       
\hline
\multicolumn{2}{c}{\ \vspace{-0.2em}}     \\ 
\multicolumn{2}{c}{Olympic dataset \cite{niebles2010modeling} \vspace{0.2em}}     \\ 
\hline   
Jain \etal \cite{jain2013better}                                       & $83.2$                 \\ \hline      
Li \etal \cite{li2013dynamic}                                         & $84.5$                \\ \hline      
Gaidon \etal \cite{gaidonIJCV2014}                              & $85.5$             \\ \hline      
Wang \etal \cite{wangIJCV2015}                                  & $90.4$        \\ \hline      
Lan \etal \cite{lan2015beyond}                                      & $\mathbf{91.4}$                 \\ \hline      
A-LOMo (proposed)                                                              & $\mathbf{91.2}$             \\       
\hline
\end{tabular}
}
\vspace{1em}
\label{tab_lomo_33}
 \scalebox{1.0}{
\renewcommand\arraystretch{1.0} 
         \begin{tabular}{|r|r|C|}
\multicolumn{3}{c}{HMDB dataset \cite{kuehneICCV2011} \vspace{0.2em}}     \\ 
\hline
Sharma \etal \cite{sharma2015action}                & LSTM + Soft-attention model (end-to-end)       & $41.3$   \\ \hline     
Gaidon \etal \cite{gaidonIJCV2014}                  & Tree Kernel + Clustered Trajectories           & $41.3$   \\ \hline     
Jain \etal \cite{jain2013better}                    & Compensated Trajectories + VLAD                & $52.1$   \\ \hline      
Simonyan \etal \cite{SimonyanNIPS2014}              & RGB(VGG)-CNN + Flow(VGG)-CNN (end-to-end)                & $59.4$   \\ \hline   
Wang \etal \cite{wangIJCV2015}                      & iDT-256-FV (pyramid)                           & $60.1$   \\ \hline      
Hoai \etal \cite{hoai2014improving}                 & iDT-256-FV (pyramid) + Relative Class Scores   & $60.8$   \\ \hline      
Zhu \etal \cite{zhu2016key}                         & Key-volume mine w/ RGB(VGG) + Flow(VGG) (end-to-end) & $63.3$   \\ \hline      
Fernando \etal \cite{fernando2015modeling}          & Rank Pooling (iDT-256-FV)                      & $63.7$   \\ \hline      
Li \etal \cite{li2016videolstm}                     & iDT-256-FV + Video-LSTM (end-to-end) + Objects & $64.9$   \\ \hline     
Lan \etal \cite{lan2015beyond}                      & iDT-256-FV + Multi-skip Feature Stacking       & $65.1$   \\ \hline      
Bilen \etal \cite{bilen2016dynamic}                 & AlexNet + Dynamic Network + IDT 				 & $65.2$   \\ \hline     
Feichtenhofer \etal \cite{FeichtenhoferCVPR2016}    & iDT-256-FV + RGB(VGG) + Flow(VGG)              & $\mathbf{69.2}$   \\ \hline     
A-LOMo (proposed)                                   & Trained on ResNet, C3D, iDT-64-FV              & $66.0$   \\ \hline  
A-LOMo (proposed) + Objects                         & Adaptive LOMo + Objects                        & $67.1$ \\     
\hline  
\end{tabular}
}
\vspace{-1em}
\caption{Comparison of the proposed approach with several state-of-the-art methods on three
human action recognition datasets. }
\vspace{-1.4em}
\label{tab_lomo_34}
\end{table*}

\subsection{Parameter Study}

\begin{figure}[t]
        \centering
        \includegraphics[width=0.47\linewidth]{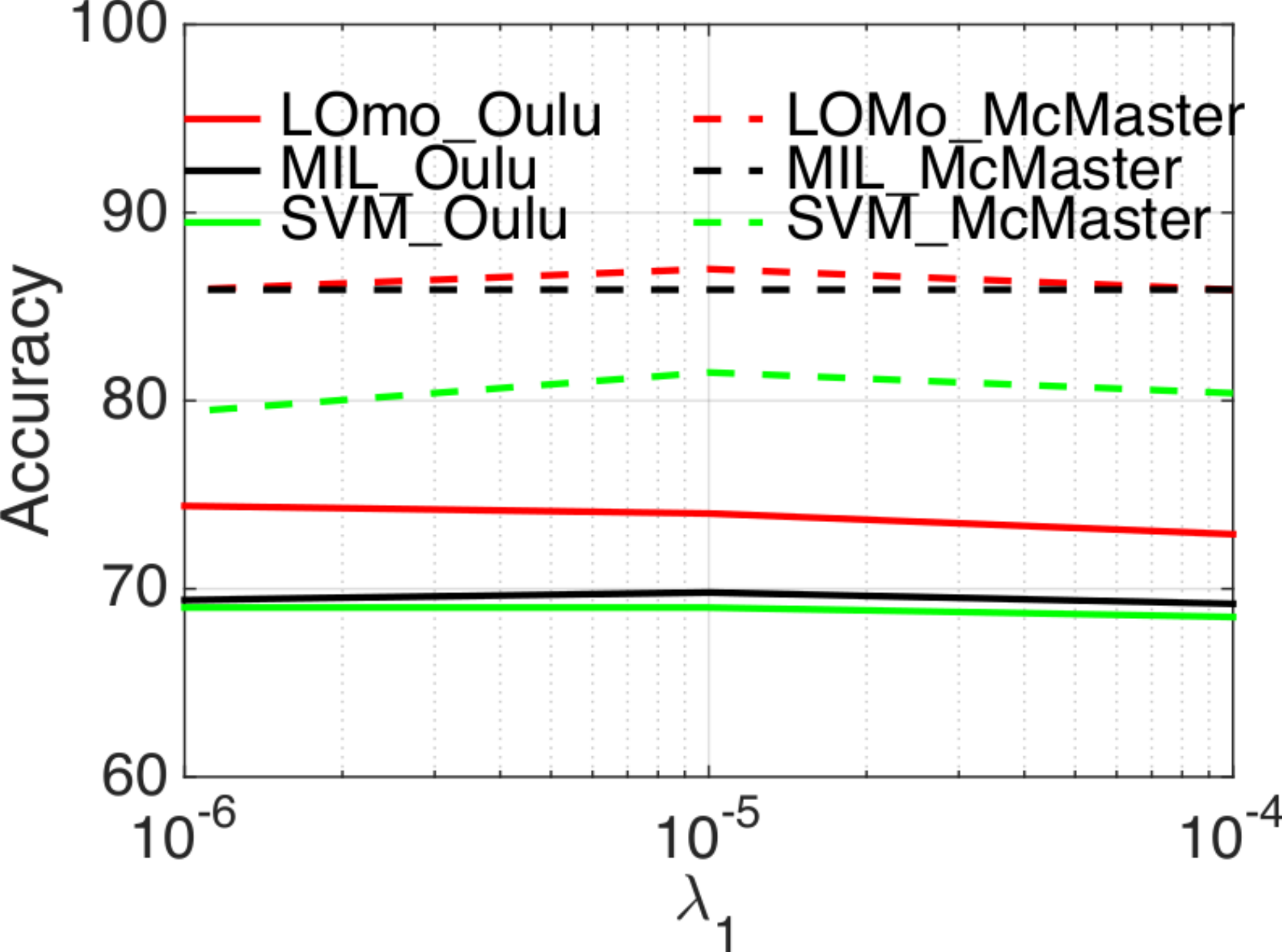} \quad 
        \includegraphics[width=0.47\linewidth ]{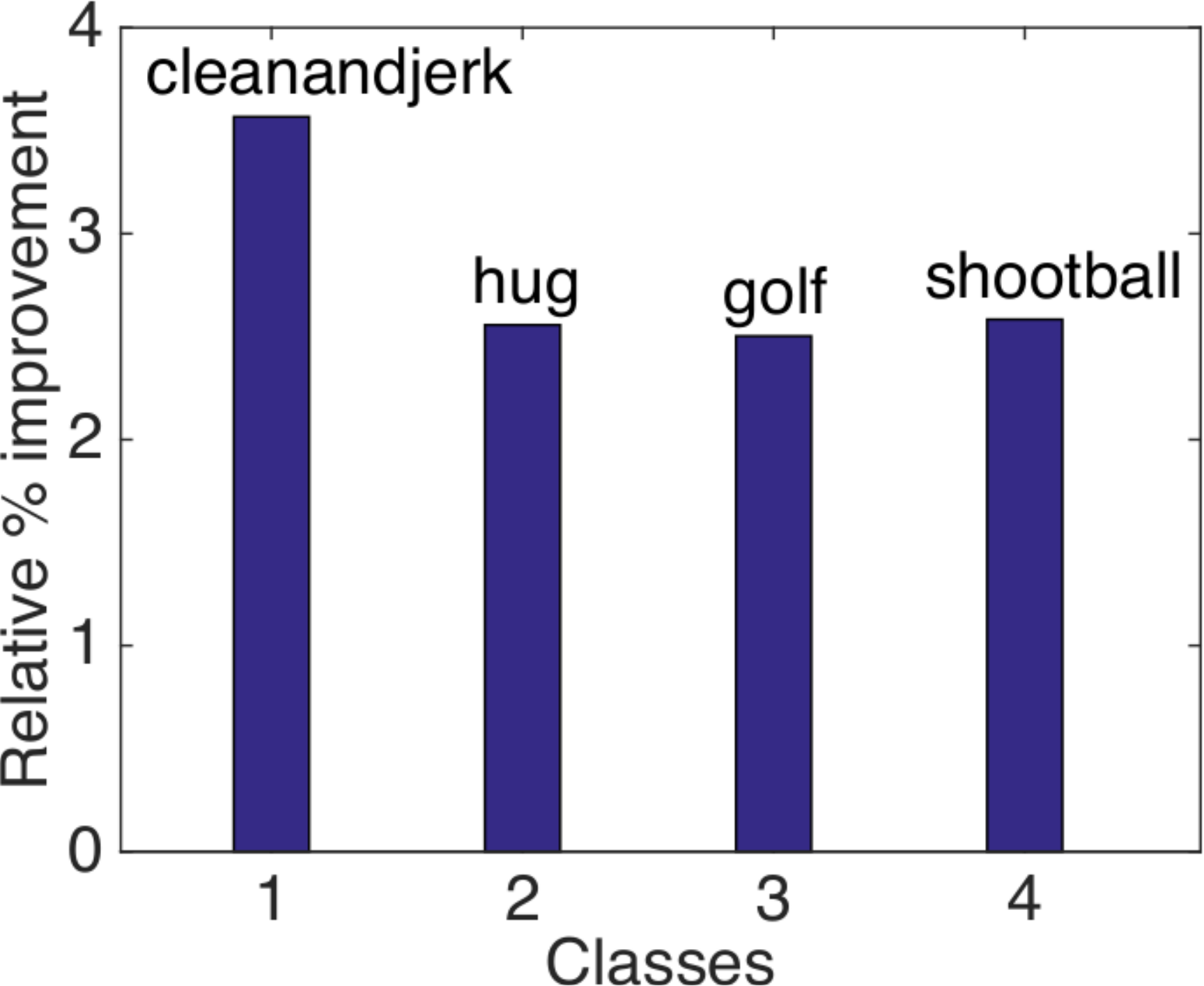} 
        \vspace{-0.8em}
        \caption{(Left) The effect of varying the regularization parameter $\lambda_1$
                and (right) the relative gain in performance by using ordinal cost in the scoring
                function. } 
        \vspace{-1em}
        \label{lambda_plot}
\end{figure}

\begin{figure}[t]
        \centering 
        \includegraphics[width=0.6\linewidth]{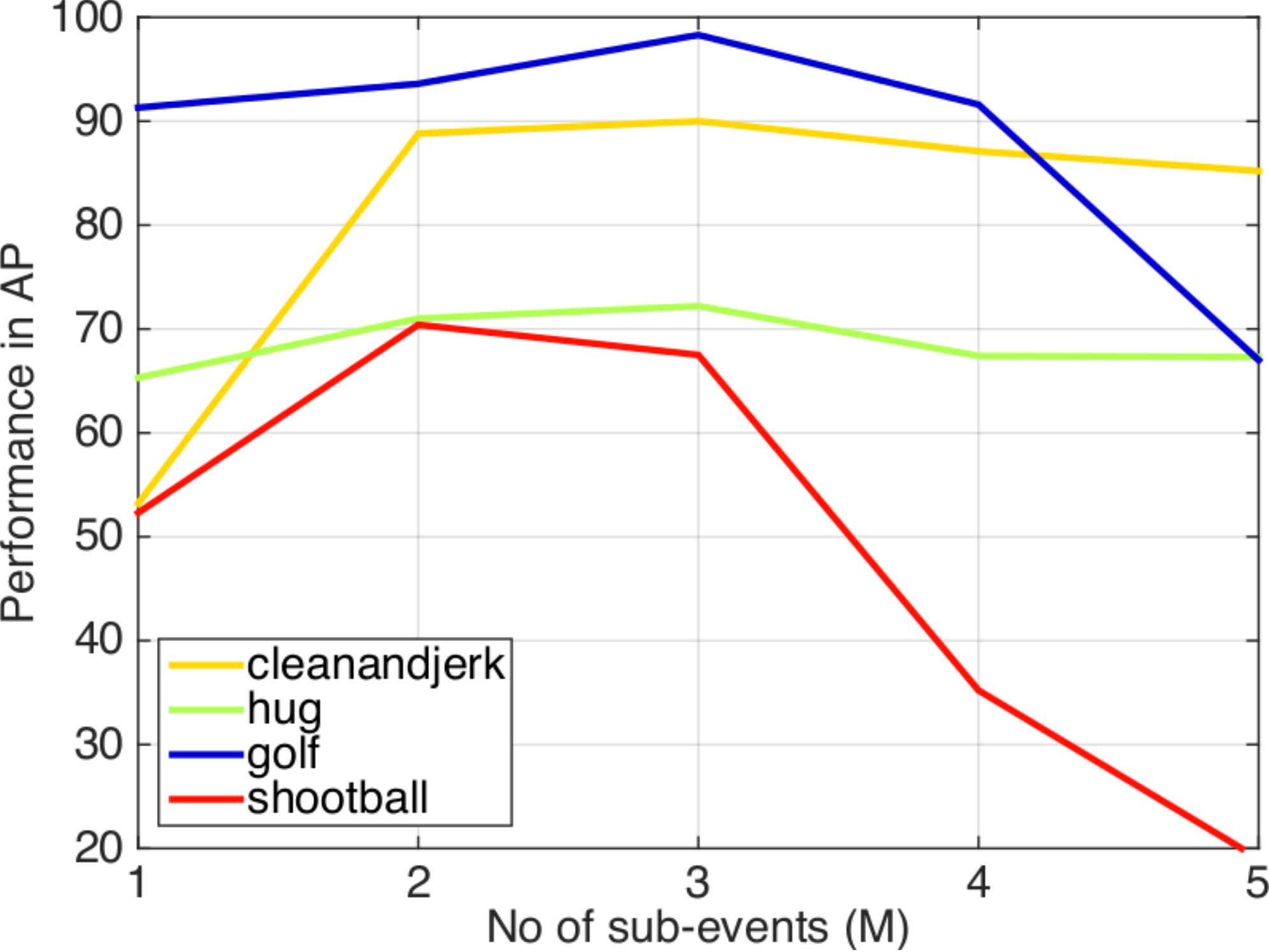} 
        \vspace{-0.8em}
        \caption{The effect of varying the number of sub-events $M$.}
        \vspace{-1.8em}
        \label{m_plot}
\end{figure}

We now study the effect of some parameters on the proposed algorithms.  \autoref{lambda_plot} (left) shows a
plot of parameter $\lambda_1$ (regularizer parameter for weights of the hyperplanes) for LOMo and our implementation of
SVM and MIL on two facial behavior datasets. We clearly see that the results are not sensitive to $\lambda_1$ and LOMo
also shows clear improvements over the baseline algorithms. In order to show the advantages of using ordinal information
inside the model, we selected the same $4$ classes (``hug'', ``cleanandjerk'', ``golf'' and ``shootball'') as used in
the qualitative results in \autoref{human_action_qual}. These classes seem to have a distinct temporal structure and are
composed of fine-grained sub-events that differ in their motion patterns. \autoref{lambda_plot} (right) shows relative
improvements for $4$ classes by switching off the learned ordinal component in the scoring function of LOMo during
testing. We observe improvements, for these classes ($2\%$ on average), while using the ordinal cost in the scoring
function. We also show a plot of mAP scores versus the number of events ($M$) used to train the model in
\autoref{m_plot}. We see that for most classes the performance goes up from $M=1$ to $M=3$ and then goes down for higher
vales of $M$. The only exception is the class ``shootball'' where results for $M=2$ are higher as compared to $M=3$, and
this is the case since two sub-events could be better at representing the class. Also with $M=2$ for the ``shooball''
class, we found that using the ordinal part of the scoring function yields a relative improvement of $2.5\%$. 

\subsection{Comparison with State-of-the-Art}
\subsubsection{Human Facial Analysis}
In this section we compare our approach with several existing approaches on the three
facial expression datasets (CK+, Oulu-CASIA VIS and UNBC McMaster). \autoref{tab_lomo_2} shows our results
along with many competing methods on these datasets. To obtain the best performance from the model,
we exploited the complementarity of different facial features by combining LOMo models learned on
three facial descriptors -- SIFT based, geometric and LBP. We combined the models with late fusion
by averaging the outputs of all the models. With this setup, we achieve state-of-the-art results on
the three datasets. 

Several initial methods pooled the spatio-temporal information in the videos, \eg (i)
LBPTOP \cite{zhao2007dynamic} -- Local Binary Patterns in three planes (XY and time), (ii) HOG3D
\cite{klaser2008spatio} -- spatio-temporal gradients, and (iii) 3D SIFT \cite{scovanner2007}. We
report results from Liu \etal \cite{liu2014learning}, who used a similar experimental protocol.
These were initial works and we see that their performances are far from current methods, \eg
compared to $81.2\%$ for the proposed LOMo, HOG3D obtains $70.6\%$ and LBPTOP obtains $72.1\%$ on
the Oulu-CASIA VIS dataset.  

Approaches modeling temporal information include Exemplar-HMMs \cite{sikkaexemplar}, STM-ExpLet
\cite{liu2014learning}, MS-MIL \cite{viola2006multiple}. While Sikka \etal (Exemplar-HMM)
\cite{sikkaexemplar} computed distances between exemplar HMM models, Liu \etal (STM-ExpLet)
\cite{liu2014learning} learned a flexible spatio-temporal model by aligning local spatio-temporal
features in an expression video with a universal Gaussian Mixture Model.  LOMo outperforms such
methods on both the emotion classification tasks, \eg on Oulu-CASIA VIS dataset, LOMo achieves a
performance improvement of $7.5\%$ and $6.5\%$ absolute relative to STM-ExpLet and Exemplar-HMMs
respectively.   Sikka \etal \cite{sikka2014classification} first extracted multiple temporal
segments and then used MIL based on boosting MIL \cite{viola2006multiple}. Chongliang \etal
\cite{Wu2015FG} extended this approach to include temporal information by adapting HMM to MIL. We
also note the performance in comparison to both MIL based approaches (MS-MIL
\cite{sikka2014classification} and MIL-HMM \cite{Wu2015FG}) on the pain dataset. Both the methods
reported very competitive performances of $83.7\%$ and $85.2\%$ on UNBC McMaster dataset compared to
$87.0\%$ obtained by the proposed LOMo.  Since having a large amount of data is difficult for many
facial analysis tasks, \eg clinical pain prediction, our results also show that combining, simple
but complementary, features with a competitive model leads to higher results.

\subsubsection{Human Analysis}

In this section, we compare our approach with several existing approaches for human activity
classification on three benchmark datasets (Olympic, HighFive and HMDB). \autoref{tab_lomo_34} shows
our results on other competing methods on these datasets. In order to obtain the best performance
from the model, we exploit the complementarity of different features by combining LOMo models
learned on three different descriptors -- iDT and deep features from C3D and ResNet-152\footnote{We
        tried both VGG-16 and ResNet-152 feautures and decided to opt for the later due to higher
        performance for these experiments.}. As the standard performance metric for HMDB dataset is
mean multiclass accuracy, we fix $\lambda_1 = 10^{-4}$ to calibrate scores from different one-vs-all
classifiers. We use late fusion for combining these features by doing weighted averaging over their
normalized prediction scores. Since different methods report results by combining multiple
approaches, for HMDB dataset we also give brief description of the corresponding methods. 

We first consider methods that relied on encoding motion and appearance information using low-level
features, followed by pooling operation to obtain a fixed length vector.
These approaches generally extracted variants of motion trajectories to encode motion, \eg Wang \etal
\cite{wangIJCV2015} extracted improved trajectories with motion stabilization, Jain \etal
\cite{jain2013better} compensated for camera motion by removing background motion, Jiang \etal
\cite{jiang2012trajectory} clustered trajectories to model high level motion patterns. We show
consistent performance improvements compared to these methods e.g. on HMDB dataset we achieve 
$66.0\%$ versus $60.1\%$ of Wang \etal, $52.1\%$ of Jain \etal and $40.7\%$ of Jiang \etal. We also
see similar improvements on the Olympic dataset ($90.4\%$ of Wang \etal vs $91.2\%$ by Adaptive
LOMo).  When compared with a recently proposed method to improve pre-existing features by stacking
them at multiple scales \cite{lan2015beyond}, we achieve similar results on Olympic ($91.2\%$
versus $91.2\%$) and better results on HMDB ($66.0\%$ versus $65.1\%$) datasets. Since our method
uses the standard iDT we believe that further improvements are possible by using such methods.

We also compare with approaches that additionally encoded temporal or ordering information inside the
classification model while training on the above features. On the Olympic dataset we achieve
$91.4\%$ mAP compared to (i) $85.5\%$ by Gaidon \etal \cite{gaidonIJCV2014} using kernels to
compare motion hierarchies in videos, (ii) $84.5\%$ by Li \etal \cite{li2013dynamic} using dynamic
pooling, and (iii) $82.0\%$ by Liu et al \cite{liu2016hierarchical} who modeled complex activities as
composed of decomposable actions.  In comparison to a recent method by  Hoai \etal
\cite{hoai2014improving}, that used distribution of classifier scores and relative class scores for
classification, we achieve $66.0\%$ mAP versus $60.8\%$ on HMDB dataset. On HighFive dataset our
performance was lower in comparison to Hoai \etal \cite{hoai2014improving} by a small margin
($70.0\%$ for LOMo versus $71.1\%$ of Hoai \etal). On HMDB dataset we also show improvement in
comparison to a recent method that used rank pooling \cite{fernando2015modeling} (instead of mean
pooling) over iDT-256-FV features for classification ($3.6\%$ relative improvement).

Several recent approaches have relied on using deep architectures for action
recognition. Simonyan \etal \cite{SimonyanNIPS2014} proposed a two-stream
convolutional network for action recognition that made predictions by late
fusion over RGB and optical flow based networks.  On HMDB dataset we achieve
$66.4\%$ mAP as compared to $59.4\%$ by this two-stream network. We also
outperform methods relying on end-to-end learning of deep features with
recurrent neural networks such as soft-attention based LSTMs
\cite{sharma2015action} ($66.0\%$ versus $41.3\%$). We also compare
our method with a recent method utilizing fully convolutional LSTM architecture
that was build upon both motion and RGB information and can both classify and
localize an action \cite{li2016videolstm}.  For a fair comparison with their
method, we also fuse the classification scores of a model that used softmax
scores from a CNN trained on $15000$ objects as descriptors, by Jain \etal
\cite{jain2015what}.  In comparison to their performance of $64.9\%$ on HMDB, we
achieve $66.0\%$ without object score fusion and $67.1\%$ with object score
fusion. This is interesting since their method learned an end-to-end
architecture from both motion and appearance information and also fuses the
score with iDT.  Feichtenhofer \etal \cite{FeichtenhoferCVPR2016} report a
slightly higher performance of $69.2\%$ with the fusion of an improved
two-stream network with iDT features. Recently Carreira and Zisserman \cite{carreira2017quo} showed
the advantages of pre-training deep models with much bigger datasets ($400$ classes with $400$ or
more videos for each class), and reported large performance improvements, \eg $80.7\%$ on HMDB
dataset. We would expect to see such improvements with the proposed models as well, by using better
pre-trained features. In this section we showed that our model yields results that are competitive
to existing methods which use similar amounts of traning data.

\section{Conclusion}
We presented a (loosely) structured latent variable model that
discovers prototypical and discriminative sub-events and learn a prior on the order in which they
occur in the video. We learned the model with a regularized max-margin hinge loss minimization which
we optimize with an efficient stochastic gradient descent based solver. We evaluated our model on
four challenging datasets of expression recognition, clinical pain prediction and intent prediction
in dyadic conversations as well as three challenging datasets for human analysis in video which
contain variety of actions, \eg sports actions, human-human interactions and human-object
interactions. We demonstrated with experimental results that the proposed model consistently
improves over other competitive baselines based on spatio-temporal pooling and Multiple Instance
Learning, while working with one type of state-of-the-art feature. Further, in combination with
complementary features, we showed that the model achieves state-of-the-art results on all the facial
analysis datasets while being competitive to the state-of-the-art on the human action recognition
datasets. We also showed qualitative results demonstrating the improved modeling capabilities of the
proposed method for both, facial and human analysis, tasks. The model is a general ordered sequence
prediction model and we aim to extend it to other sequence prediction tasks. Further, the classifier
learning is decoupled from the features and given the recent success of representation learning
methods, we would explore end-to-end learning of the features and classifier as another future
direction.

\ \\

\noindent
\textbf{Acknowledgements.} The authors thank Sanjoy Dasgupta 
and Harpreet Sawhney for discussions.
Gaurav Sharma was
supported by the Early Career Research Award from SERB India (ECR/2016/001158) and the
Research-I foundation at IIT Kanpur. Karan Sikka was supported by NIH grant R01 NR013500. Any
opinions, findings, conclusions or recommendations expressed in this material are those of the
author(s) and do not necessarily reflect the views of the National Institute of Health.

\ifCLASSOPTIONcaptionsoff
  \newpage
\fi
\bibliographystyle{IEEEtran}
\bibliography{egbib}

\begin{IEEEbiography} [{\includegraphics[width=1in,height=1.25in,trim=50 50 30 100,clip,keepaspectratio]{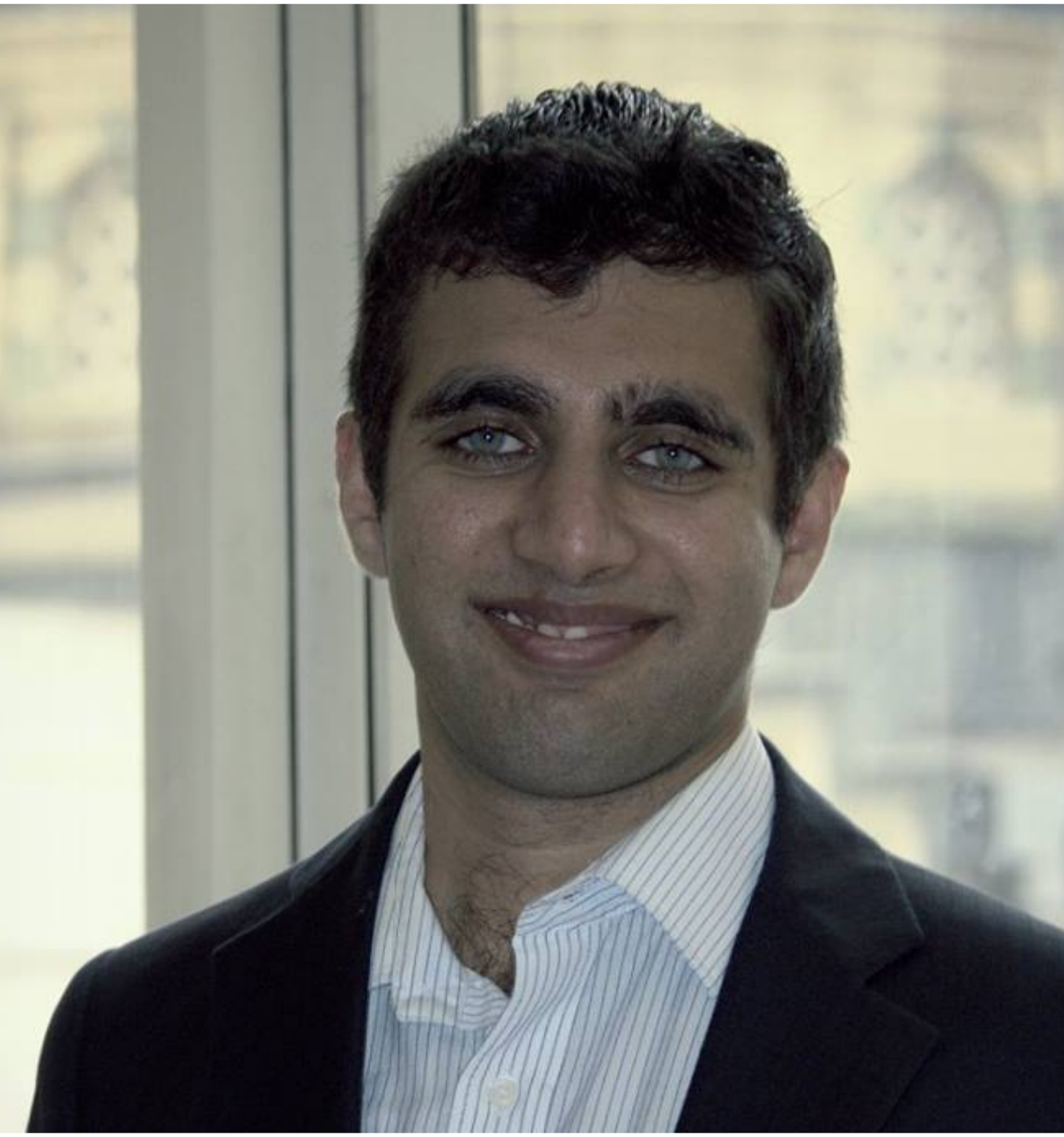}}]{Karan Sikka} 
is with SRI International, Princeton. 
%
He holds a Masters degree in ECE from UCSD and a B.Tech degree
in ECE from Indian Institute of Technology Guwahati (IIT Guwahati).  His primary research interest
lies in Computer Vision and Machine learning. 
\end{IEEEbiography}
\vspace{-3em}
\begin{IEEEbiography} [{\includegraphics[width=1in,height=1.25in,clip,keepaspectratio]{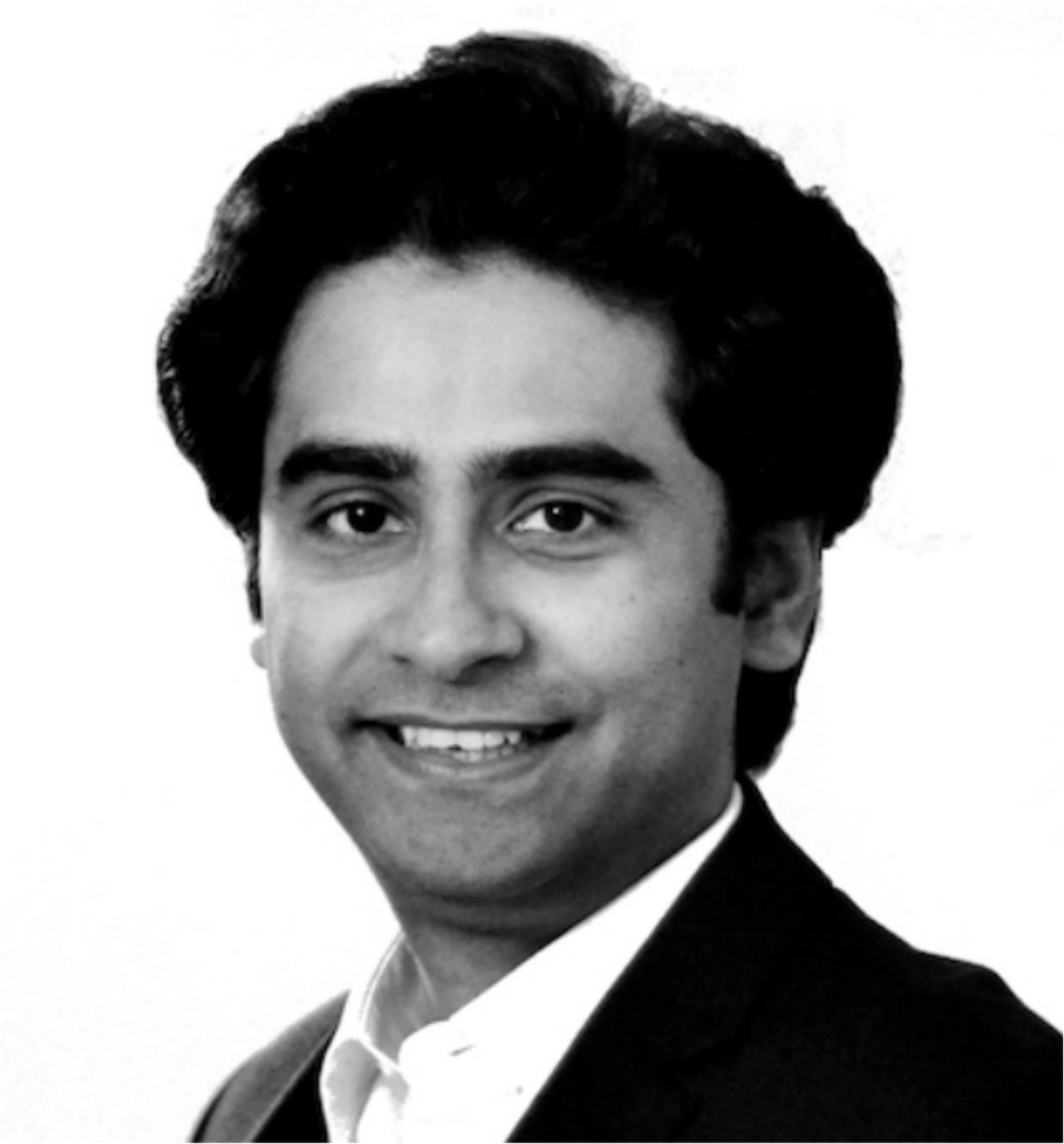}}]{Gaurav Sharma}
is with the Indian Institute of Technology Kanpur. He holds an Integrated Masters in Mathematics and
Computing from the Indian Institute of Technology Delhi and a PhD in Applied Computer Science from
INRIA (team LEAR, now THOTH) and the CNRS-GREYC, University of Caen, France.  In his previous roles
he has spent time with Samsung R\&D, Technicolor R\&I and Max Planck Institute for Informatics. His
primary research area is Computer Vision and Machine Learning. 
\end{IEEEbiography}

\onecolumn
\section{Appendix}
\subsection{Dynamic Programming for Inference}

The scoring optimization can be solved exactly using Dynamic Programming (DP).
However, in practice we found the DP based solver to be slow and resorted to
an approximate but much faster algorithm (see Sec.~3.4). The time complexity of the greedy
solution is $\mathcal{O}(NDK + NK)$ while that of exact DP solution is
$\mathcal{O}(NDK  + N^2KK!)$, where $N$ is the number of frames, $K$ is number
of events, and $D$ is dimensionality of the features. While the greedy solution
is clearly a very coarse approximation of the true scoring objective, we found
it to be very fast and well performing in the experiments and hence preferred
it over the exact dynamic programming based solver.
We compared the performance of the greedy and DP based solution on the Olympic
dataset and found the running time of the DP based solution to be $\sim 5
\times$ of greedy solution, while giving comparable final classification
performance. 

\subsection{Effect of Initialization on LOMo}
We studied the effect of initialization on LOMo by evaluating it on the
Olympic dataset with iDT features with $10$ different random seeds. The mean and standard deviation
the performance was $84.7$ and $0.7$ respectively. This mean performance is comparable to the
performance of $84.6$ reported in Tab.~2 and the standard deviation does not change any comments on
the comparison with any baseline method.

\subsection{Visual Examples from Facial Analysis Task}

\begin{figure}
	\centering
	\includegraphics[width=0.33\columnwidth,trim=0 0 0 
	0, clip]{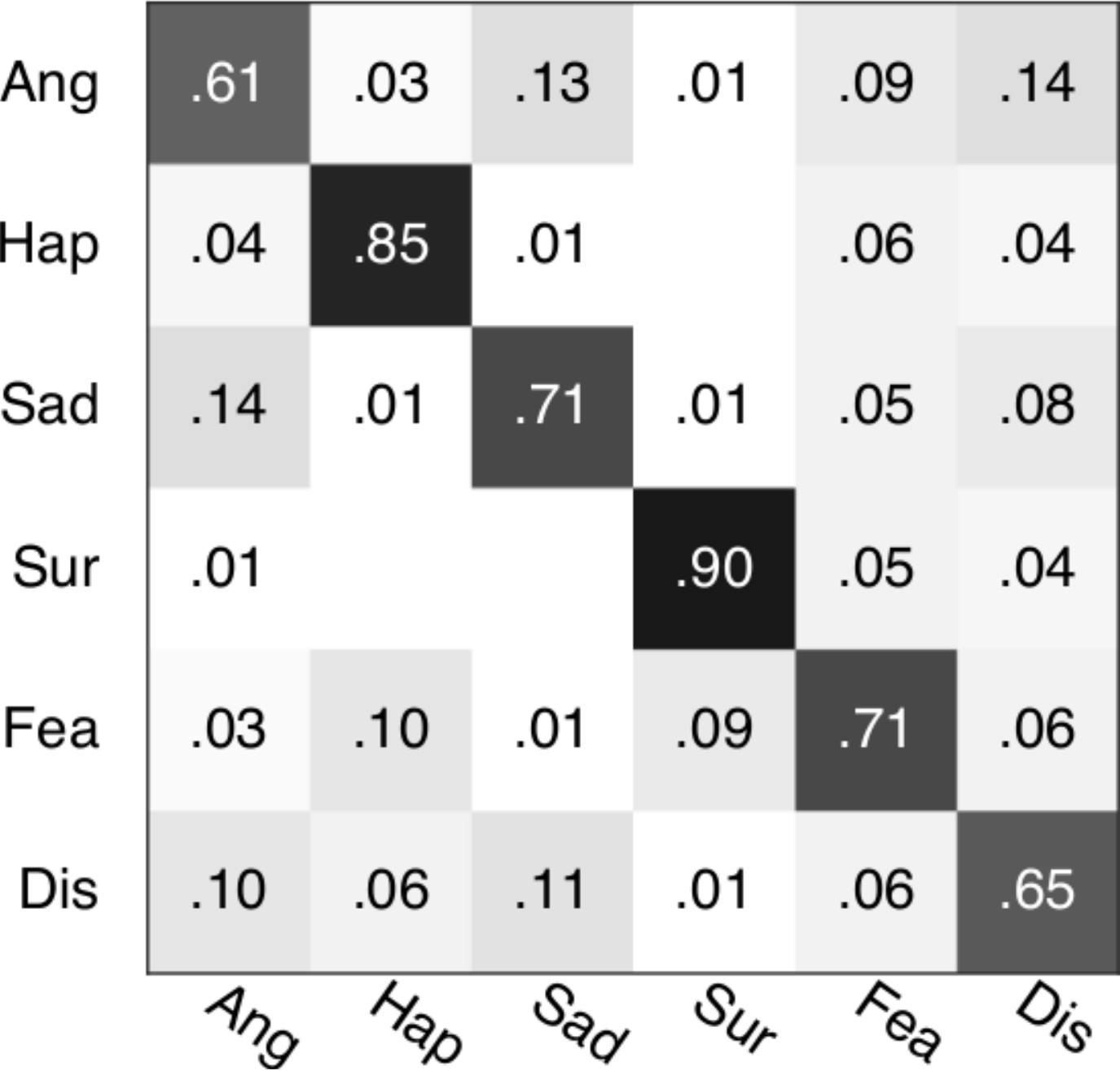} \hspace{3em}
	\includegraphics[width=0.33\columnwidth,trim=0 0 0 
	0, clip]{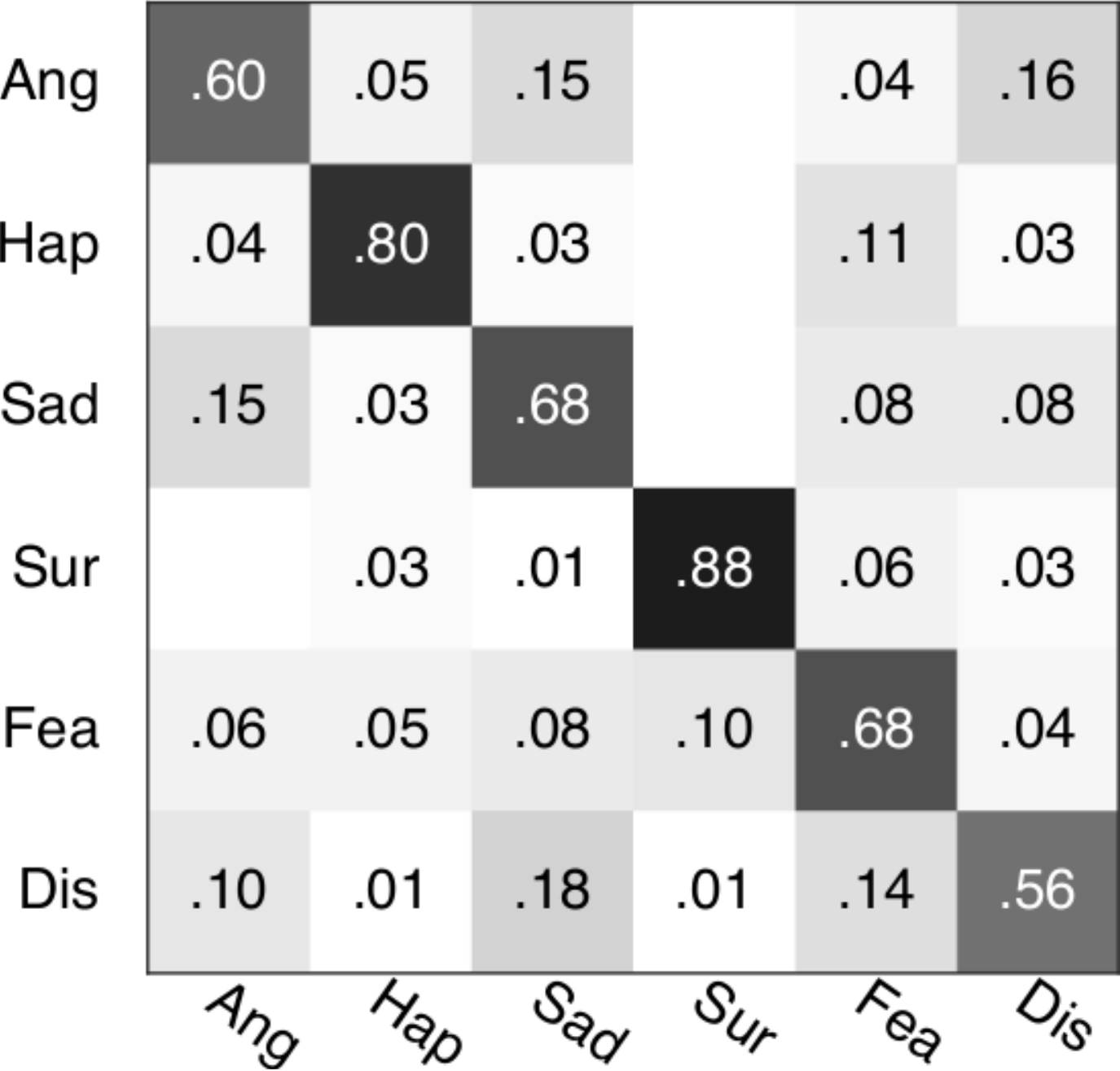}
	\caption{Confusion matrices for LOMo (left) and
	MIL (right)}
	\label{conf}
\end{figure}

\begin{figure}
	\centering
	\includegraphics[width=0.7\columnwidth,trim=180 360
	1300 200,clip]{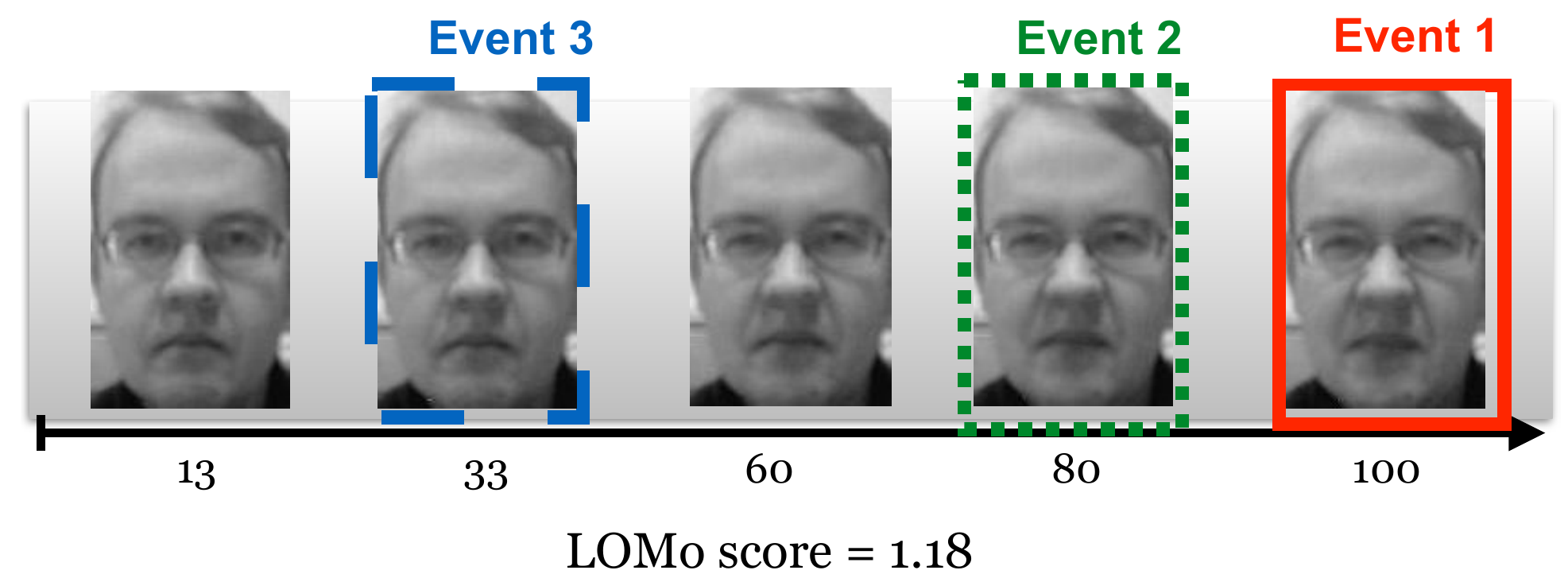} \\
	\includegraphics[width=0.7\columnwidth,trim=180 360
	1300 200,clip]{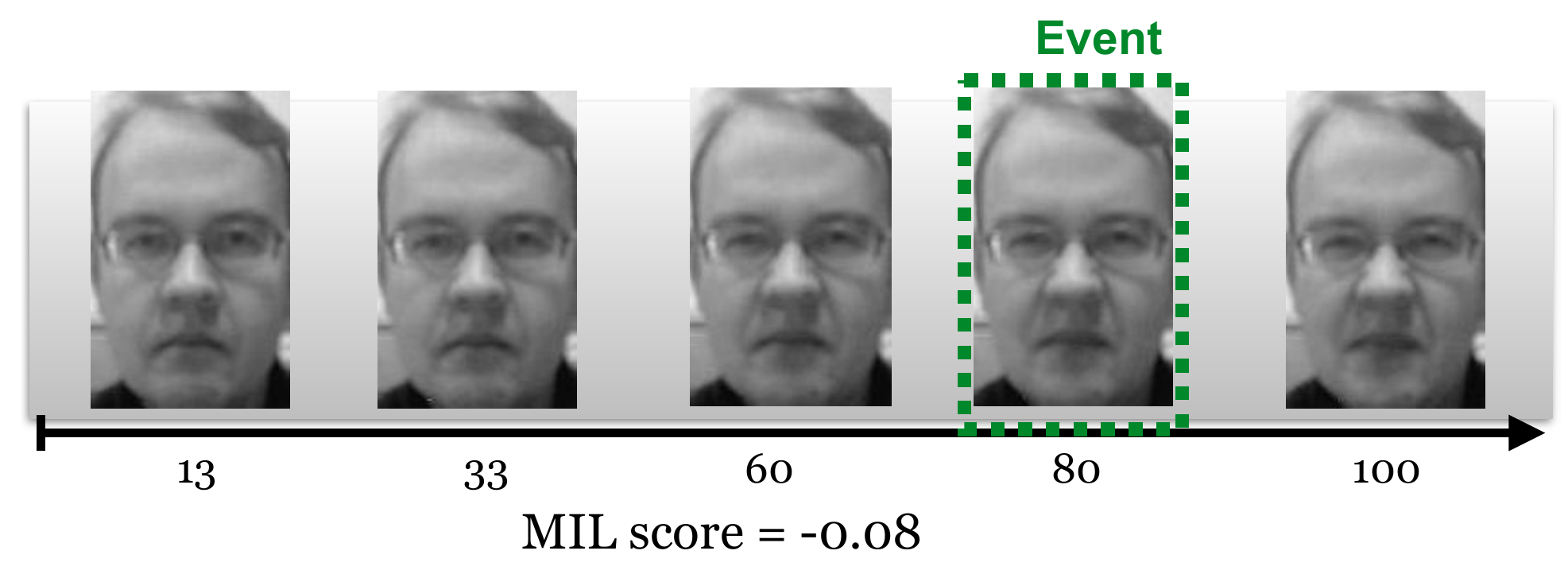} \\
	\caption{Example showing detection for
		individual sub-events with LOMo (top)
		and MIL (bottom) for an
		example from `disgust' class. The classification
		score obtained by LOMo and MIL are shown on the
	bottom of each figure. LOMo is able to learn a better classification model by learning sub-events and a prior on their ordering for the `disgust' class.}
	\label{oulu_dis}
\end{figure}

In order to support our arguments regarding the results in Sec.~4.3.1 we first
show the confusion matrices for MIL and LOMo on Oulu dataset in \autoref{conf}.
Based on the individual class performances we observe that LOMo shows substantial 
improvements in comparison to MIL on classes such as `disgust', `happy', and
'fear'. In particular the improvement on `disgust' class (which is also the
most confusing class) is significant ($\sim 9\%$ absolute). We believe this improvement results
from the ability of LOMo to capture discriminative sub-events specific to
the `disgust' class, which makes it easy to discriminate samples from this class from visually similar classes (`sad' and `fear'). We have shown
an example from the `disgust' class in \autoref{oulu_dis} along with detections
for individual sub-events with LOMo and MIL. The score obtained by LOMo (1.18)
is higher compared to the score obtained by MIL (-0.08).  We attribute this to
the ability of LOMo to both detect multiple sub-events and to model
prior on their ordering (see Sec.~4.4).  

\begin{figure}
	\centering
	\includegraphics[width=0.7\columnwidth,trim=180 385 1300 200,clip]{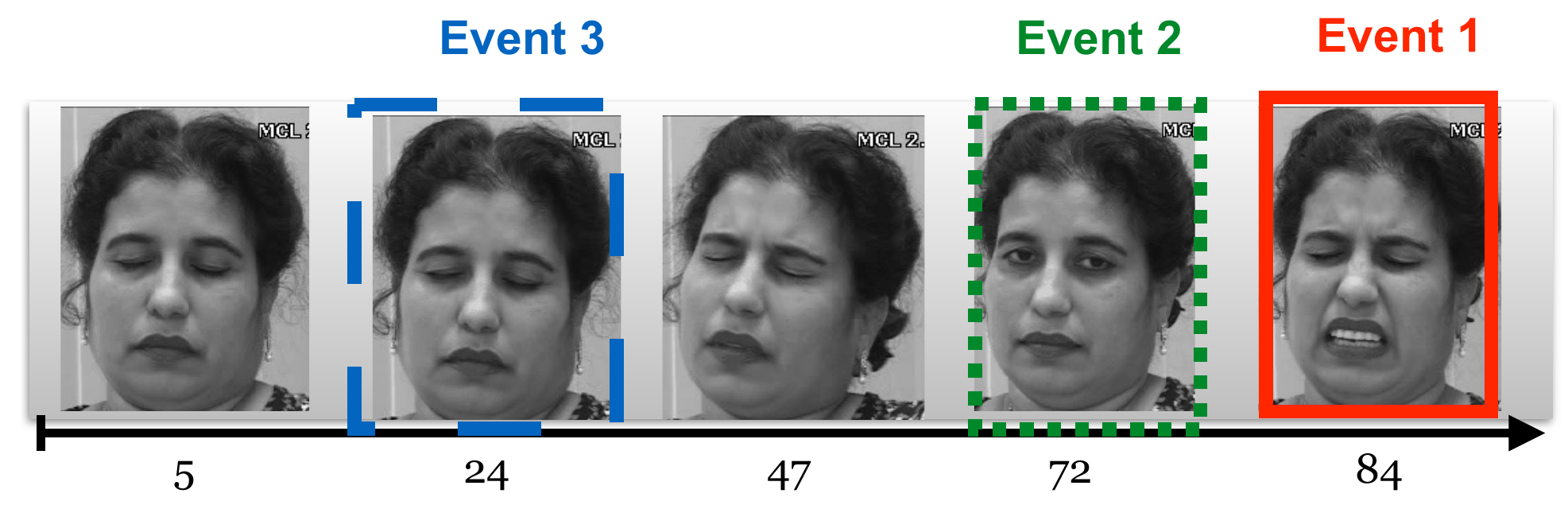} 
	\caption{Detection of multiple discriminative sub-events, discovered by LOMo, on a video
		sequence from the UNBC McMaster Pain dataset. The number below the timeline shows the relative
	location (in percentile of total number of frames).} \label{figs_mcmaster}
\end{figure}

\autoref{figs_mcmaster} shows detections on an example sequence from the UNBC McMaster dataset
where subjects show multiple expressions of pain. The results show that our approach is able to detect such multiple expressions
of pain as sub-events. 

\begin{figure}
\centering
\includegraphics[width=0.8\columnwidth]{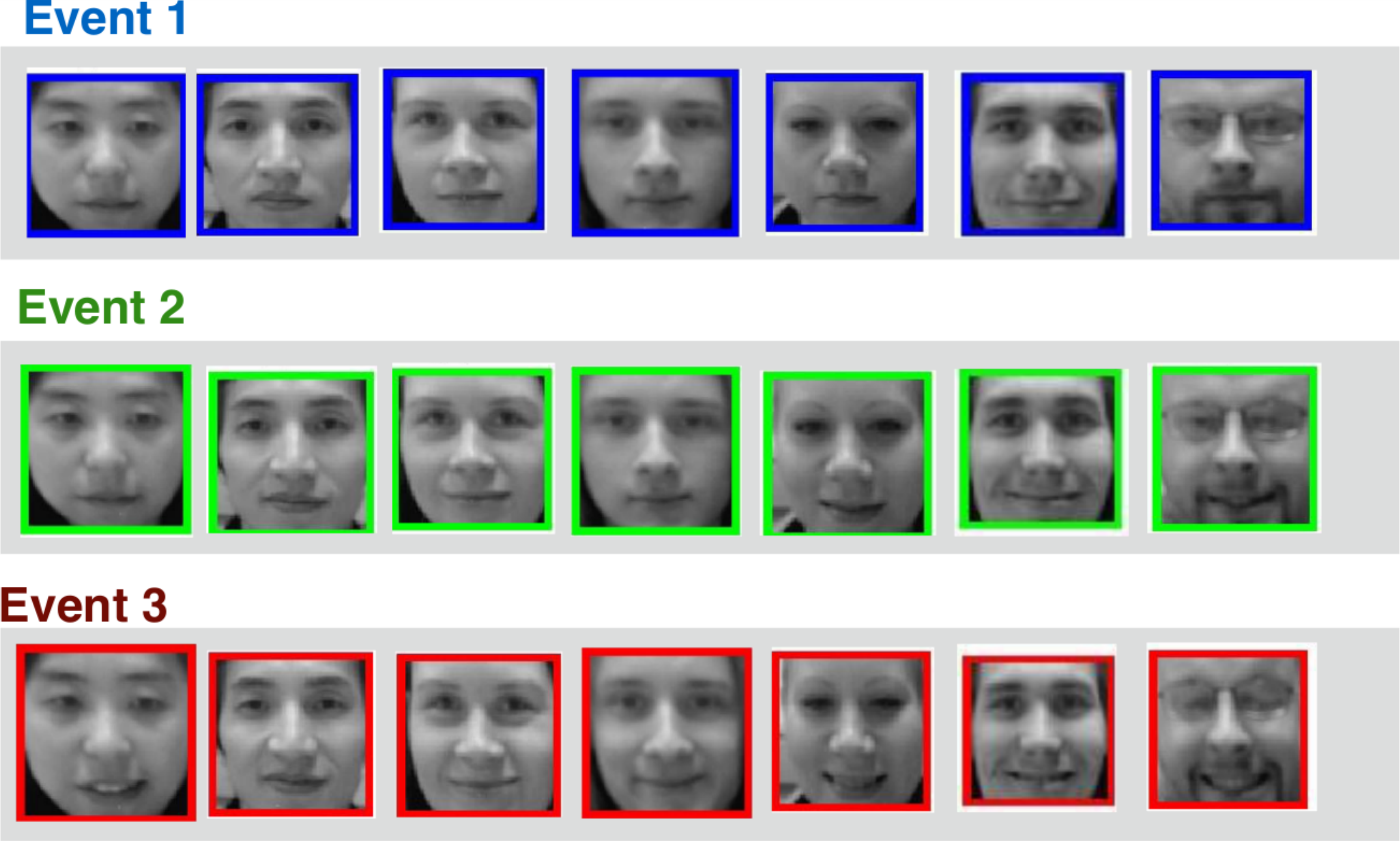}
\caption{Frames corresponding to latent sub-events as identified by our
algorithm on different subjects. This figure shows results for LOMo trained for
classifying `happy' expression and tested on unseen test samples belonging to the
`happy' class.} 
\label{fig_plot2}
\end{figure}

For better understanding the model, we show the frames corresponding to each
latent sub-event as identified by LOMo across different subjects. Ideally each
sub-event should correspond to a facial state and thus have a common structure
across different subjects. As shown in \autoref{fig_plot2}, we see a common
semantic pattern across detected events, with `happy' classifier,  where Event 1 seems to be similar
to neutral, Event 2 to onset and Event 3 to apex phases of facial expression formation. We observed
similar qualitative results across other expression classes.

\end{document}